\documentclass{new_tlp}
\usepackage{times,helvet,courier}

\usepackage{url}

\usepackage{pdflscape}
\usepackage{enumitem}
\usepackage{multienum}
\usepackage[english]{babel}
\usepackage[utf8]{inputenc}
\usepackage{amsmath}
\usepackage{graphicx}
\usepackage[colorinlistoftodos]{todonotes}
\usepackage{framed}
\colorlet{shadecolor}{gray!50}
\usepackage{fancyhdr}
\usepackage{graphicx}
\usepackage[percent]{overpic}
\usepackage{xcolor}
\usepackage{comment}
\usepackage{subfigure}
\usepackage{float}
\usepackage{booktabs}
\usepackage{changepage}
\usepackage{caption}

\usepackage{listings}
\usepackage{xcolor}
\newtheorem{Example}{Example}

\newcommand{\MathWord}[1]{\mbox{\it #1}}
\renewcommand{\MathWord}[1]{\text{\textit{#1}}}
\DeclareGraphicsExtensions{.jpg, .png, .eps}

\title[Manipulation of Articulated Objects using Dual-arm Robots via ASP]{Manipulation of Articulated Objects using Dual-arm Robots via Answer Set Programming}

\author[R. Bertolucci, et al.] {
RICCARDO BERTOLUCCI\\
University of Calabria, Italy\\
\email{bertolucci@mat.unical.it}
\and
ALESSIO CAPITANELLI\\
Teseo srl, Italy\\
\email{alessio.capitanelli@teseotech.com}
\and
CARMINE DODARO, NICOLA LEONE\\
University of Calabria, Italy\\
\email{\{dodaro,leone\}@mat.unical.it}
\and
MARCO MARATEA, FULVIO MASTROGIOVANNI\\
University of Genoa, Italy\\
\email{\{marco.maratea,fulvio.mastrogiovanni\}@unige.it}
\and 
MAURO VALLATI\\
University of Huddersfield, UK\\
\email{m.vallati@hud.ac.uk}
}

\jdate{March 2020}
\pubyear{2020}
\pagerange{\pageref{firstpage}--\pageref{lastpage}}
\doi{S1471068401001193}

\begin{document}

\label{firstpage}
\maketitle

\begin{abstract}

The manipulation of articulated objects is of primary importance in Robotics, and can be considered as one of the most complex manipulation tasks. 
Traditionally, this problem has been tackled by developing ad-hoc approaches, which lack flexibility and portability.

In this paper we present a framework based on Answer Set Programming (ASP) for the automated manipulation of articulated objects in a robot control architecture.  
In particular, ASP is employed for representing the configuration of the articulated object, for checking the consistency of such representation in the knowledge base, and for generating the sequence of manipulation actions. 

The framework is exemplified and validated on the Baxter dual-arm manipulator in a first, simple scenario. 
Then, we extend such scenario to improve the overall setup accuracy, and to introduce a few constraints in robot actions execution to enforce their feasibility. 
The extended scenario entails a high number of possible actions that can be fruitfully combined together.
Therefore, we exploit macro actions from automated planning in order to provide more effective plans.
We validate the overall framework in the extended scenario, thereby confirming the applicability of ASP also in more realistic Robotics settings, and showing the usefulness of macro actions for the robot-based manipulation of articulated objects.
Under consideration in Theory and Practice of Logic Programming (TPLP).
\end{abstract}


\section{Introduction}
\label{sec:intro}

The manipulation of articulated objects plays an important role in real-world robot tasks, both in home and industrial environments \cite{Krugeretal2009,DBLP:conf/iros/Heyer10}. 
Much attention has been paid to the development of approaches and algorithms for generating the sequence of movements a robot has to perform in order to manipulate an articulated object. 
In the literature, the problem of determining the two-dimensional (2D) configuration of articulated or flexible objects has been vastly considered in the past few years \cite{DBLP:journals/ijrr/WakamatsuAH06,DBLP:conf/aiia/CapitanelliMMV17,DBLP:conf/icra/NairCAIAML17,DBLP:journals/ras/CapitanelliMMV18}, whereas the problem of obtaining a target object configuration via manipulation has been explored in robot motion planning \cite{DBLP:conf/isrr/SchulmanHLA13,Yamakawaetal2013,DBLP:journals/tase/BodenhagenFJWAOKPK14}. 
A limitation of such manipulation strategies is that they are often crafted specifically for the problem at hand, with the relevant characteristics of the object and robot capabilities being either hard coded or assumed a-priori known.
Therefore, in these approaches generalisation properties and scalability are somehow limited. 
In this paper we present a framework based on Answer Set Programming (ASP)~\cite{DBLP:journals/amai/Niemela99,baral_2003,DBLP:journals/cacm/BrewkaET11} for the automated manipulation of articulated objects in a robot 2D workspace. 
ASP is a general, prominent knowledge representation and reasoning language with roots in logic programming and non-monotonic reasoning~\cite{DBLP:conf/iclp/GelfondL88,DBLP:conf/iclp/GelfondL91}. 
In particular, in this paper ASP is employed for representing the configuration of the articulated object, for checking the consistency of such representation in the knowledge base, as well as for generating the sequence of manipulation actions, i.e., the plan. 

The framework is first validated using a Baxter dual-arm manipulator in a \textit{simple scenario}, which involves the manipulation of an articulated object in a 2D workspace with the possibility of performing such actions like rotating one of its link with respect to another one around their joint. 
Afterwards, we extend the simple scenario to allow for a higher accuracy of the mockup, and we introduce some constraints in robot actions execution to enforce their feasibility. 
Such \textit{extended scenario} entails a high number of actions to be possibly executed in sequence, which can be fruitfully combined.
Therefore, we exploit \textit{macro} actions \cite{ChrpaVM15,GereviniSV15,ChrpaV19} from automated planning in order to generate more compact and effective plans. 
Macro actions can be considered as sequences of \textit{elementary} actions that, on an application viewpoint, would be useful to be performed in such a sequence, and be considered as a \textit{single} action.
In Robotics applications, macro actions may be useful given that performing a sequence of multiple elementary -- yet atomic -- actions is expected to be more time consuming than executing the relative macro.
In particular, in our application scenario, we have defined macro actions considering sequences of actions that are \textit{often} performed in sequence by the robot, thus meeting such desirable property.
We validate the whole framework in this extended scenario, confirming
the applicability of ASP-based knowledge representation and reasoning in more realistic Robotics contexts as well, at the same time showing the usefulness of macro actions in this peculiar application. 

This paper is an extended and revised version of a paper appearing in the proceedings of LPNMR'19 \cite{DBLP:conf/lpnmr/BertolucciCDLMM19}. The main improvements of the current paper are: $(i)$ the definition of a new, more realistic scenario, $(ii)$ the extension and validation of the whole framework on this new scenario, and $(iii)$ the exploitation of macro actions for the new scenario, including a new Action Planning encoding employing macro actions, and related experimental analysis of the module and validation of the framework.

The paper is structured as follows. 
Section~\ref{sec:assumptions} presents the problem statement and the simple  scenario. 
Section~\ref{sec:arch} shows the overall robot control architecture, while Section~\ref{sec:modu} details the modules where ASP is employed. 
The framework validation in the simple scenario is discussed in Section~\ref{sec:vali}. Section~\ref{sec:ext} introduces the extended scenario, as well as the needed extensions to the ASP encoding.
Section~\ref{sec:macro} presents the defined macros, and how they have been encoded in ASP. 
The paper ends in Section~\ref{sec:exp} by presenting the analysis we have done on the extended scenario, by discussing related research in Section~\ref{sec:rela}, and by drawing some conclusions in Section~\ref{sec:conc}.


\section{Problem Statement and the Simple Scenario}
\label{sec:assumptions}

In this Section we define more in detail the addressed problem, and we introduce the simple scenario we consider first.

\subsection{Problem Statement}

\begin{figure}[t!]
\centering
\includegraphics[width=85mm]{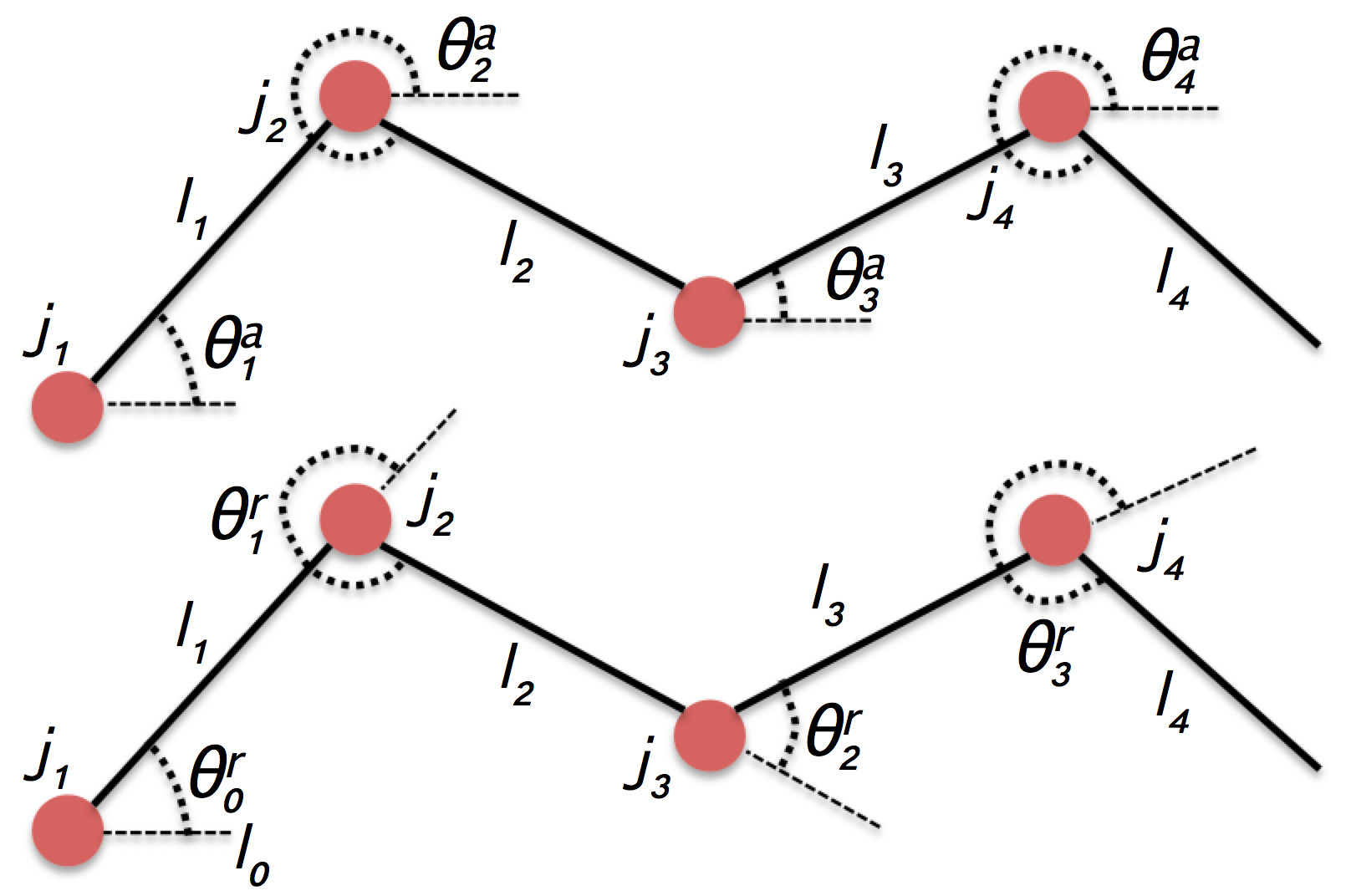}
\caption{Two possible object representations: absolute (top) and relative (bottom).}
\label{fig:representation}
\end{figure}

Our goal is to present
$(i)$ an efficient ASP-based representation, planning and execution architecture for the manipulation of articulated objects in terms of perceptual features, their representation and the planning of manipulation actions, which maximises the likelihood of being successfully executed by dual-arm robot manipulators, and
$(ii)$ given a specific articulated object's goal configuration, determine a plan to attain it, whereby each step involves one or more manipulation actions to be executed by a dual-arm robot.
In so doing, our working assumptions are as follows:
\begin{enumerate}
\item[$A_1$] flexible objects can be appropriately modelled as articulated objects with a high number of links and joints, as it is customary \cite{Yamakawaetal2013};
\item[$A_2$] an articulated object is manipulated while placed on a table, and therefore its 2D configuration is only affected by robot manipulation actions, whereas the effects of external forces such as gravity are not considered; 
\item[$A_3$] we do not consider possible issues related to grasping or motion dexterity during the manipulation task in reasoning process;
\item[$A_4$] sensing is affected by noise, but the \textit{symbol grounding problem}, i.e., the association between perceptual features and the corresponding symbols \cite{Harnad1990}, is assumed to be solved.
\end{enumerate}

On the basis of assumption $A_1$, we focus on articulated objects only. 
We define an articulated object as a pair $\alpha = \left(\mathcal{L}, \mathcal{J}\right)$, where $\mathcal{L}$ is the ordered set of its $|\mathcal{L}|$ links and $\mathcal{J}$ is the ordered set of its $|\mathcal{J}|$ joints.  
Each link $l \in \mathcal{L}$ is characterised by two parameters, namely a length $\lambda_l$ and an orientation $\theta_l$. 
Considering assumption $A_2$, we restrict the set of all possible configurations to 2D configurations only. 
Also, we allow only for a limited number of possible orientations, which induces a finite set of allowed angle values. 
If $\alpha$ is represented using \textit{absolute} angles (i.e., with respect to an external Cartesian reference frame as shown in Figure \ref{fig:representation} on the top), then its configuration is an $|\mathcal{L}|$-tuple:
\begin{equation}
C_{\alpha, a} = \left(\theta^a_1, \ldots, \theta^a_{|\mathcal{L}|}\right). 
\label{eq:absolute}
\end{equation}
Otherwise, if relative angles are used (e.g., with respect to the previous link, as shown in Figure \ref{fig:representation} on the bottom), then the configuration must be \textit{augmented} with an initial \textit{hidden} link $l_0$ in order to define a reference frame needed for actual manipulation actions:
\begin{equation}
C_{\alpha, r} = \left(\theta^r_{0}, \theta^r_1, \ldots, \theta^r_{|\mathcal{L}|} \right).
\label{eq:relative}
\end{equation}
In fact, while in principle the relative angles approach could represent the configuration of an articulated object with one joint less compared with the absolute angles one, the resulting representation would not be unique (actually, there would infinitely many), since the object maintains relative orientations among its parts even when rotated \textit{as a whole}.
Obviously enough, this is not a practical way for grounding manipulation actions in the robot workspace, while at the representation level would provide nonetheless a complete representation of the articulated object configuration \cite{DBLP:journals/ras/CapitanelliMMV18}.
In this paper, we adopt an absolute angles representation, and we refer to the configuration of an articulated object $\alpha$ simply as $C_{\alpha}$.

The problem we want to formalise and solve in this paper can be described as follows.
Given an articulated object $\alpha$, and having defined the number of its $|\mathcal{L}|$ links and $|\mathcal{J}|$ joints, we want to determine an ordered sequence of grounded actions, i.e., a plan $\mathcal{P}$, i.e.,
\begin{equation}
\mathcal{P} = \left\{a_1, \ldots, a_{|\mathcal{P}|}\right\},    
\end{equation}
\noindent such that its execution by a dual-arm robot manipulator leads from an initial articulated object configuration $C_{\alpha, i}$ to a goal configuration $C_{\alpha, g}$, whereas actions in $\mathcal{P}$ are instances of a-priori defined planning operators.
The set of planning operators differs whether we consider the simple or the extended scenario. 
Whilst in the simple scenario we model only manipulation actions aimed at re-orienting pairwise links, in the extended scenario we also model link grasping and releasing operators by robot grippers.

\begin{figure}[t!]
\centering
\includegraphics[width=85mm]{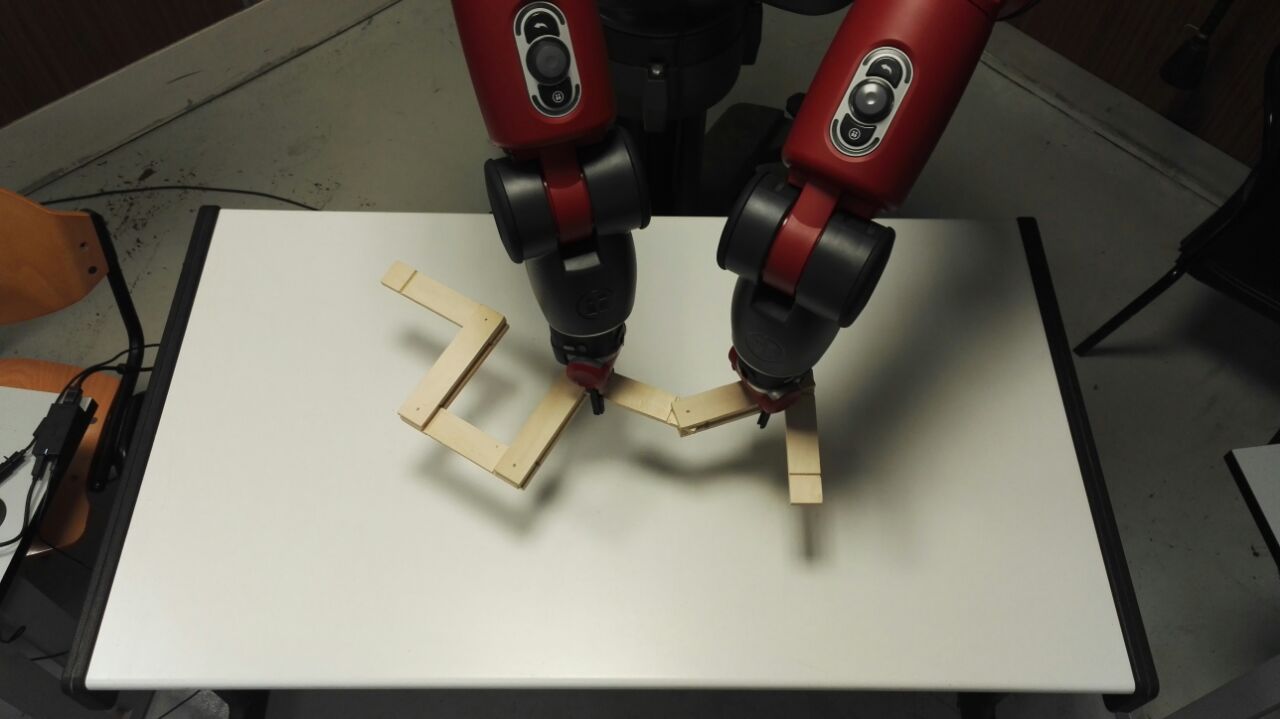}
\caption{The experimental scenario.}
\label{fig:scenario}
\end{figure}

\subsection{Simple Scenario}
\label{label:srs}

In order to comply with assumption $A_2$, which allows us to focus on the manipulation process, we have setup a scenario in which a dual-arm Baxter robot manipulates an articulated object that is conveniently located on a table in front of it.

In the setup, the table sustains the articulated object while it is being manipulated by the robot, and it is assumed to be large enough to accommodate the whole object itself, see Figure \ref{fig:scenario}.
As a consequence, link rotations occur only around axes centred on specific object joints, but always perpendicular to the table surface.
We have crafted two wooden articulated objects of different size: 
the first, which is simpler, has three $40$ $cm$ long links (which are connected by two in-between joints), whereas the second is made up of five $20$ $cm$ long links (connected by four\ joints).
For both objects, links are $2$ $cm$ wide and $3$ $cm$ thick.
The two objects have been designed to reduce the likelihood of manipulation-specific issues when using the Baxter's standard grippers, in order to comply with assumption $A_3$.
The Baxter's \textit{head} is equipped with a camera pointing downward to the table and able to acquire images of the relevant robot manipulation workspace.
QR tags are attached to each object link, which reduces perception errors and is aimed at enforcing assumption $A_4$.
Each QR code provides an overall 6D link pose, which directly maps to an absolute link orientation $\theta^a_l$.

\begin{figure}[t!]
\centering
\includegraphics[width = 120mm]{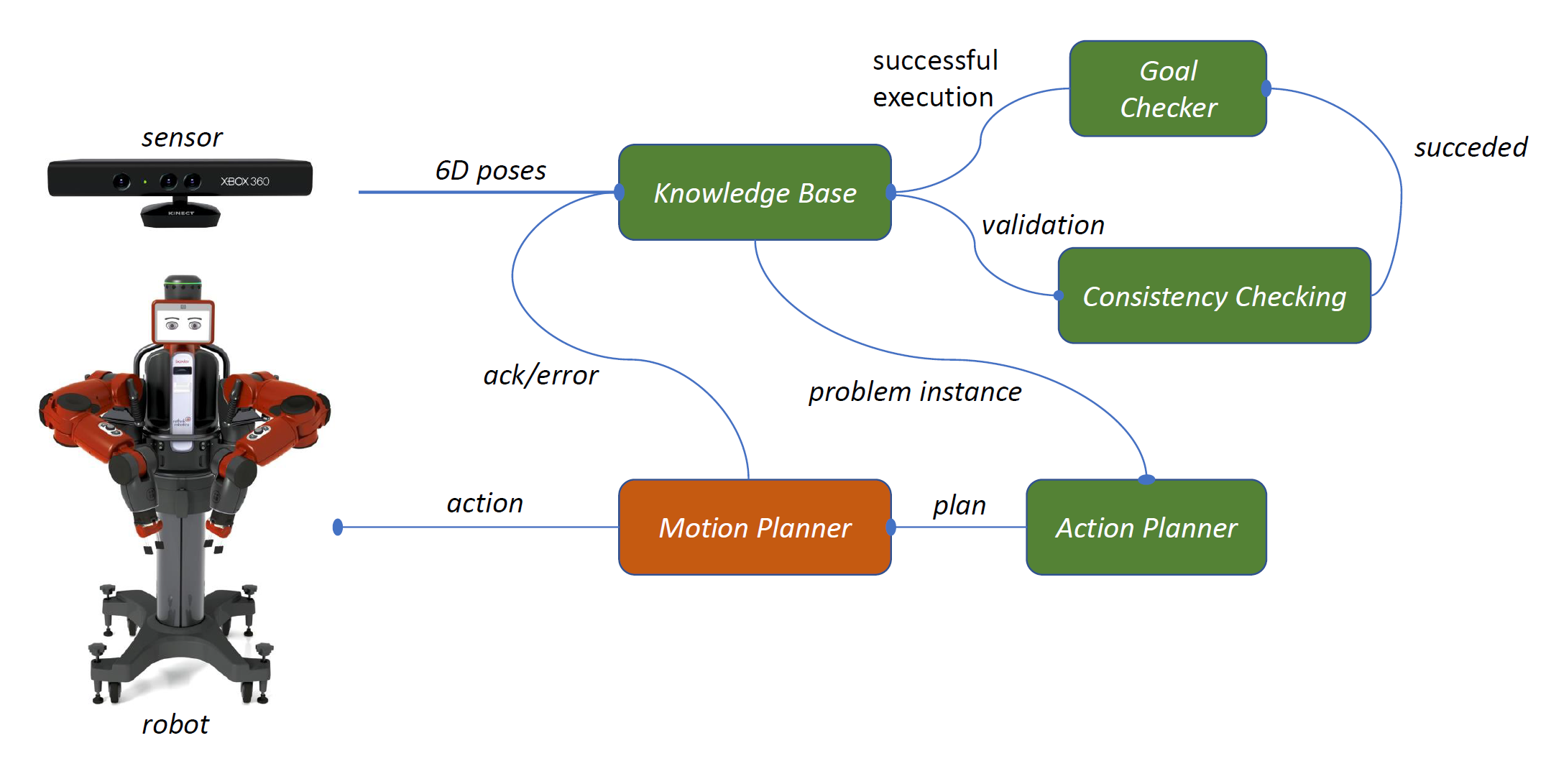}
\caption{The system's architecture: in \textit{green} ASP-based modules, in \textit{orange} the robot-specific control module.}
\label{fig:architecture}
\end{figure}

\section{System's Architecture}
\label{sec:arch}

Stemming from previous work \cite{DBLP:conf/aiia/CapitanelliMMV17,DBLP:journals/ras/CapitanelliMMV18}, the overall system's architecture is a hybrid, reactive/deliberative framework including perceptual, knowledge representation and reasoning, as well as action modules, as shown in Figure \ref{fig:architecture}.
Although the target robot platform is the Baxter dual-arm robot manipulator from Rethink Robotics, in principle the architecture can be adapted to other manipulators as well, either in simulation or in real-world conditions, as long as appropriate perception, low-level motion planning algorithms, and manipulation strategies are adopted. 

In the current implementation, perception modules acquire a video stream using an RGB-D camera sensor located on top of the robot's \textit{head} and pointing downward, which provides 6D poses for each link.
In order to simplify and make the perception process more robust, the various links of the articulated object are uniquely identified by fiduciary markers, i.e., QR tags.
Although more advanced perception strategies are certainly possible, this does not limit the results and the scope of this paper. 
The perception pipeline generates a set of 6D pose data structures, each one for an object link, which are used to update corresponding ASP-based representation structures in the \textit{Knowledge Base} module. 
Once such representation structures in \textit{Knowledge Base} are updated, the \textit{Consistency Checking} module verifies that facts within it are not mutually inconsistent, and that numerical values associated to joint angles are withing the proper ranges.
Relevant parts of the current ASP-based knowledge base are processed to this aim by the rules encoded in the \textit{Consistency Checking} module.
As a result, a ASP-based problem instance is generated, whose goal and initial conditions depend on the target articulated object configuration, and the current configuration maintained in the \textit{Knowledge Base} module, respectively.
The \textit{Action Planner} module receives such problem instance and generates a plan in the form of an ordered sequence of manipulation actions to be performed by the robot on the articulated object.
Once a plan is generated, its actions are processed sequentially to drive the overall behaviour of the robot by the \textit{Motion Planner} module, 
which is responsible for the execution of all manipulation actions.
Each action in the simple scenario involves rotating a target link with respect to another, connected one, and in particular around axes centred on the link joints, whose poses are maintained in the \textit{Knowledge Base} module. 
Any action may be either successful or not, depending on a number of reasons related to perception noise, and grasping or manipulation faults in real-world environments.
If a manipulation action is successful, the \textit{Motion Planner} proceeds with the one that follows until the plan ends and the \textit{Knowledge Base} module is notified about successful execution.
Otherwise, an issue is raised and re-planning occurs, thereby reiterating the whole work-flow described above.
The \textit{Goal Checker} is indeed aimed at detecting whether the remaining part of the (already computed) plan can be successfully executed.
If not, a re-planning process must occur using the current perceived environment configuration as a new initial configuration.

It is noteworthy that all modules except \textit{Motion Planner} (i.e., all \textit{green} modules in Figure \ref{fig:architecture}) are based on ASP. 
The \textit{Motion Planner} module encodes all the robot-specific details related to kinematics constraints, dynamics, as well as control.
The interested reader is referred to \cite{Darvishetal2018} for a thorough description of the module.
One of our architectural assumptions is the separation between the deliberative layer (implemented with ASP encodings) and the robot control layer, although such an assumption is characterised by well-known limitations, discussed in \cite{Darvishetal2018}. 
As such, we aimed at abstracting and simplifying as much as possible all the issues typically related to more realistic scenarios in Robotics, i.e., perception, grasping, and manipulation. 
To this aim, we decided not to consider the articulated object physical properties, such as material or friction, and we assume its perception to be \textit{perfect} since we use QR tags to identify its links. 
Grasping and manipulation do not employ any feedback-in-the-loop sensing, and unsuccessful actions are recognised only by their effects detected after actions end by the expected location of QR codes.



\section{ASP-based Modules in the System's Architecture}
\label{sec:modu}
In this Section we describe how ASP is used to implement the modules depicted 
in Figure~\ref{fig:architecture}. 
In the following, we assume that the reader is familiar with ASP and ASP-Core-2 input language specifications~\cite{CalimeriFGIKKLM20}. 


\subsection{Knowledge Base}
\label{sec:kb}

The knowledge base consists of facts over atoms of the form \texttt{joint(J)}, \texttt{isLinked(J1,J2)}, \texttt{angle(A)},  \texttt{time(T)}, \texttt{hasAngle(J,A,T)}, and \texttt{goal(J,A)}, and the constants \texttt{timemax} and \texttt{granularity}.
Atoms over the predicate \texttt{joint} represent the joints of the articulated object.
Atoms over the predicate \texttt{isLinked} represent links between joints \texttt{J1} and \texttt{J2}.
Atoms over the predicate \texttt{angle} represent the possible angle values that can be taken by joints, and they can range from $0$ to $359$. 
Actually, the atom \texttt{angle(0)} must be always part of the knowledge base and admissible angles are the ones that can be obtained by rotating a joint by the degrees specified by the constant \texttt{granularity}, e.g., if the granularity is $90$ degrees, then the admissible angles are $0$, $90$, $180$, and $270$.
Atoms over the predicate \texttt{time} represent the possible time steps, and they range from $0$, which represents the initial state, to \texttt{timemax}. 
Atoms over the predicate \texttt{hasAngle} represent the angle \texttt{A} of the joint \texttt{J} at time \texttt{T}. 
The knowledge base only contains the initial state of each joint, i.e., its angle at time $0$.
Finally, atoms over the predicate \texttt{goal} represent the angle \texttt{A} that must be reached by the joint \texttt{J} at the time step specified by \texttt{timemax}.
An example of the input is represented by the facts and constants reported in Figure~\ref{fig:valid-asp1}. 
It is noteworthy that the constant \texttt{timemax} is not included in the example, and its usage will be described in Section~\ref{sec:action}.
\begin{figure}[t!]
$$
\begin{array}{l}
\texttt{joint(1..5).}\quad
\texttt{angle(0).}\quad
\texttt{angle(90).}\quad
\texttt{angle(180).}\quad
\texttt{angle(270).}\\
\texttt{isLinked(1,2).}\quad
\texttt{isLinked(2,3).}\quad
\texttt{isLinked(3,4).}\quad
\texttt{isLinked(4,5).}\\
\texttt{hasAngle(1,90,0).}\quad
\texttt{hasAngle(2,180,0).}\quad
\texttt{hasAngle(3,180,0).}\\
\texttt{hasAngle(4,270,0).}\quad
\texttt{hasAngle(5,270,0).}\quad 
\texttt{time(0..timemax).}\\
\texttt{goal(1,270).}\quad
\texttt{goal(2,270).}\quad
\texttt{goal(3,180).}\quad
\texttt{goal(4,270).}\\
\texttt{goal(5,270).}\quad
\texttt{\#const granularity = 90.}
\end{array}
$$
\caption{An example of a knowledge base encoded in ASP for a $5$-link articulated object.
}
\label{fig:valid-asp1}
\end{figure}


\begin{figure}[t!]
    \centering
    $$
\begin{array}{ll}
c_{1a}\ & \texttt{\footnotesize{:- isLinked(J1,J2), not joint(J1).}}\\
c_{1b}\ & \texttt{\footnotesize{:- isLinked(J1,J2), not joint(J2).}}\\
c_{2}\  & \texttt{\footnotesize{:- isLinked(J,J).}}\\
c_{3a}\ & \texttt{\footnotesize{:- hasAngle(J,A,T), not joint(J).}}\\
c_{3b}\ & \texttt{\footnotesize{:- hasAngle(J,A,T), not angle(A).}}\\
c_{3c}\ & \texttt{\footnotesize{:- hasAngle(J,A,T), not time(T).}}\\
c_{4a}\ & \texttt{\footnotesize{:- goal(J,A), not joint(J).}}\\
c_{4b}\ & \texttt{\footnotesize{:- goal(J,A), not angle(A).}}\\
c_{5}\ & \texttt{\footnotesize{moreThanOneGoal(J) :- joint(J), \#count\{A:goal(J,A)\}>1.}}\\
c_{6}\ & \texttt{\footnotesize{:- joint(J), moreThanOneGoal(J).}}\\
c_{7}\ & \texttt{\footnotesize{oneStartingAngle(J) :- joint(J), \#count\{A:hasAngle(J,A,0)\}=1.}}\\
c_{8}\ & \texttt{\footnotesize{:- joint(J), not oneStartingAngle(J).}}\\
c_{9}\ & \texttt{\footnotesize{:- not time(0).}}\\
c_{10}\ & \texttt{\footnotesize{:- not angle(0).}}\\
c_{11}\ & \texttt{\footnotesize{possibleAngle(0).}}\\
c_{12}\ & \texttt{\footnotesize{possibleAngle(X) :- possibleAngle(Y), X=(Y+granularity)$\setminus$360.}}\\
c_{13}\ & \texttt{\footnotesize{:- not angle(X), possibleAngle(X).}}\\
c_{14}\ & \texttt{\footnotesize{:- angle(X), not possibleAngle(X).}}\\
\end{array}
$$
    \caption{ASP encoding for consistency checking.}
    \label{fig:consistencychecking}
\end{figure}

\subsection{Consistency Checking}
The \textit{Consistency Checking} module verifies the mutual consistency of the facts maintained within the knowledge base, in order to assure that the later action planning does not draw conclusions on inconsistent pairwise facts. 

This is done using the ASP encoding reported in Figure~\ref{fig:consistencychecking}. 
In particular, rules $c_{1a}$ and $c_{1b}$ check whether atoms over the predicate \texttt{isLinked} represent the links between two joints, while $c_2$ checks whether there is no link between the same joint.
Rules $c_{3a}$, $c_{3b}$, and $c_{3c}$ check whether the predicate \texttt{hasAngle} expresses angles in the proper format, whereas $c_{4a}$ and $c_{4b}$ check the correctness of the predicate \texttt{goal}.
Rules $c_5$ and $c_6$ check whether at most one goal is specified for each joint (in order to avoid multiple goals for the same joint which would confuse the reasoner), whereas
rules $c_7$ and $c_8$ verify whether each joint is in exactly one angle at time step 0 (for the same reason).
Rules $c_9$ and $c_{10}$ simply verify the existence of the first time step and angle 0, respectively.
Finally, rules from $c_{11 }$ to $c_{14}$ check whether atoms over the predicate \texttt{angle} represent allowed angles.



\subsection{Action Planning}
\label{sec:action}

ASP is not a planning-specific language, but it can be also used to specify encodings for planning domains~\cite{DBLP:journals/ai/Lifschitz02}, like the problem we consider in this paper.
We have defined several encoding variants, for what concerns either the manipulation modes and the strategy for computing plans. 
The encoding described in this Section is our base encoding, and it is embedded into a classical iterative deepening approach in the spirit of SAT-based planning~\cite{DBLP:conf/ecai/KautzS92}, where \texttt{timemax} is initially set to $1$ and then increased by $1$ if a plan is not found, which guarantees to generate the shortest possible plans for a sequential encoding, i.e., when the robot performs only one action for each step (see Section \ref{sec:rela} for more details about other strategies). 

\begin{figure}[t!]
$$
\begin{array}{ll}
r_{1}\ & \texttt{\footnotesize{joint(0).}}\\
r_{2}\ & \texttt{\footnotesize{hasAngle(0,0,0).}}\\
r_{3}\ & \texttt{\footnotesize{isLinked(0,1).}}\\
r_{4}\ & \texttt{\footnotesize{isLinked(J1,J2) :- isLinked(J2,J1).}}\\
r_{5}\ & \texttt{\footnotesize{\{changeAngle(J1,J2,A,Ai,T) : joint(J1), joint(J2), J1>J2, angle(A),}}\\
\hspace{10pt} & \hspace{20pt} \texttt{\footnotesize{hasAngle(J1,Ai,T), A<>Ai, isLinked(J1,J2)\} <= 1}}\\
\hspace{10pt} & \hspace{40pt} \texttt{\footnotesize{:- time(T), T < timemax, T > 0}}.\\
r_{6}\ & \texttt{\footnotesize{ok(J1,J2,A,Ai,T)  :- changeAngle(J1,J2,A,Ai,T), }}\\
\hspace{10pt} & \hspace{20pt}  \texttt{\footnotesize{F1=(A+granularity)$\setminus$360, F2=(Ai$\setminus$360), F1=F2, A < Ai.}}\\
r_{7}\ & \texttt{\footnotesize{ok(J1,J2,A,Ai,T)  :- changeAngle(J1,J2,A,Ai,T), }}\\
\hspace{10pt} &  \hspace{20pt} \texttt{\footnotesize{F1=(Ai+granularity)$\setminus$360, F2=(A$\setminus$360), F1=F2, A > Ai.}}\\
r_{8}\ & \texttt{\footnotesize{ok(J1,J2,A,0,T) :- changeAngle(J1,J2,A,0,T), A=360-granularity.}}\\
r_{9}\ & \texttt{\footnotesize{ok(J1,J2,0,A,T) :- changeAngle(J1,J2,0,A,T), A=360-granularity.}}\\
r_{10}\ & \texttt{\footnotesize{:- changeAngle(J1,J2,A,Ai,T), not ok(J1,J2,A,Ai,T).}}\\
r_{11}\ & \texttt{\footnotesize{affected(J1,An,Ac,T) :- changeAngle(J2,\_,A,Ap,T), hasAngle(J1,Ac,T), }}\\
\hspace{10pt} & \hspace{20pt} \texttt{\footnotesize{ J1>J2, angle(An), An=|(Ac + (A-Ap)) + 360|$\setminus$360, time(T).}}\\
r_{12}\ & \texttt{\footnotesize{hasAngle(J1,A,T+1) :- changeAngle(J1,\_,A,\_,T).}}\\
r_{13}\ & \texttt{\footnotesize{hasAngle(J1,A,T+1) :- affected(J1,A,\_,T).}}\\
r_{14}\ & \texttt{\footnotesize{hasAngle(J1,A,T+1) :- hasAngle(J1,A,T), not changeAngle(J1,\_,\_,\_,T),}}\\
\hspace{10pt} & \hspace{20pt} \texttt{\footnotesize{not affected(J1,\_,\_,T), T <= timemax.}}\\
r_{15}\ & \texttt{\footnotesize{:- goal(J,A), not hasAngle(J,A,timemax).}}
\end{array}
$$
\caption{Encoding for the simple scenario: it allows for forward manipulation actions only.}
\label{fig:standard_f}
\end{figure}

Figure \ref{fig:standard_f} reports the base encoding. 
It is noteworthy that it uses operations \texttt{\textbackslash} and $\lvert \cdots \lvert$, which are not defined in the ASP-Core-2 standard but supported by Clingo~\cite{DBLP:conf/iclp/GebserKKOSW16}, that compute the remainder of the division and the absolute value, respectively. 
This encoding allows for forward propagation only, i.e., manipulation actions \textit{in one direction} along the articulated object chain, e.g., it is possible to move the second joint holding the first joint but not the opposite.

Since we employ an absolute representation, $r_{1}$, $r_{2}$ and $r_{3}$ add to the knowledge base the \texttt{joint(0)}, its angle and the link to joint $1$. 
This joint is not meant at being  moved, and it is used only to have a fixed reference between the robot-centred and the articulated object frames. 
Rule $r_{4}$ enforces bi-directionality of linked joints, i.e., if \texttt{joint(1)} is linked to \texttt{joint(2}) then \texttt{joint(2)} is also linked to \texttt{joint(1)}.
Then, rule $r_{5}$ is used to select an atom of the form \texttt{changeAngle(J1,J2,A,Ai,T)}, where \texttt{J1} is the joint around which rotation occurs, \texttt{J2} is the joint related to the link to be kept steady, \texttt{A} is the desired angle, \texttt{Ai} is the current angle of \texttt{J1} and \texttt{T} is the current step.
Rule $r_{10}$ ensures the validity of the configuration represented by the atom \texttt{changeAngle(J1,J2,A,Ai,T)}, that is when each action has a desired angle \texttt{A} that can be reached in one step (rules $r_{6}$, $r_{7}$, $r_{8}$, and $r_{9}$).
Rule $r_{10}$ takes advantage of the atoms over the predicate \texttt{ok}, which are used to model the valid configurations of the atom \texttt{changeAngle(J1,J2,A,Ai,T)}.
Rule $r_{11}$ is used to identify which joints are affected by the atom selected in $r_{5}$. 
In particular, atoms of the form \texttt{affected(J1,An,Ac,T)} represent the fact that, at time step \texttt{T}, the angle associated with joint \texttt{J1} is updated from \texttt{Ac} to \texttt{An}. 
Rules $r_{12}$ and $r_{13}$ are used to update the joints' angle values for the next step, while $r_{14}$ states that if neither $r_{12}$ nor $r_{13}$ have affected any joint then the corresponding angle values remain unchanged. 
Finally, $r_{15}$ states the the goal must be reached.

It is noteworthy that the encodings presented in this paper assume that the resulting (intermediate) configurations of the articulated object are feasible in practice, e.g., in case of link overlaps. 
While it could be certainly possible to reason about the pairwise mutual position and orientation of each link, that is definitely out of the scope of this work.
It must be noted, however, that such assumption is quite common in hybrid reactive/deliberate robot control architectures, although a few works in the literature explore the interplay between the reactive and the deliberative layers \cite{Thomasetal2018,Thomasetal2019}.

\subsection{Goal Checker}
\label{sec:gc}

During the execution of a plan, it may happen that due to errors in robot manipulation actions, or as a result of human interventions in human-robot cooperation scenarios~\cite{DBLP:journals/ras/CapitanelliMMV18}, actual joint angle values (and therefore the relative displacement of links) may be different with respect to those represented in the knowledge base, which are propagated forward assuming that proper action execution.

If this happens, it is necessary to check whether the current configuration of the articulated object is still compatible with the plan as it is being executed by the robot.
While small variations could be easy to accommodate, and not requiring re-planning should the symbolic representation be preserved notwithstanding the differences in numerical values, big variations may lead to qualitatively different configurations for the articulated object, thereby the computation of a new plan from a new configuration may be needed. 
This can be accomplished by asynchronously creating a new initial object configuration based on the current robot perception about its workspace and the articulated object in particular. 
In this process, the role of the \textit{Goal Checker} module is to check whether there is the need to create such a new object configuration, i.e., when the intermediate goal have not been reached for the reasons outlined above.
This is done by using rule $r_{15}$ from the encoding in Figure \ref{fig:standard_f}.


\section{Simple Scenario and a first Experimental Validation}
\label{sec:vali}
\noindent
In order to perform a first validation of the whole architecture, we have prepared a setup whereby a dual-arm Baxter robot has to manipulate a $5$-link articulated object.
The object is located on a table in front of the robot.
Articulated objects made up of $5$ links provide a very valuable ground for testing our approach, as the \textit{limited} number of links is expected not to make the manipulation difficult for the robot, whereas such number proves to be sufficiently high to require the development of complex plans to reach a goal configuration. 
The use of a dual-arm manipulator like Baxter is justified by its widespread adoption as a research platform, and by the necessity to employ a robot with two arms in order to manipulate the object, i.e., the robot must hold one link of the object while rotating an adjacent one. 
It may be argued that robot simulations may be characterised by some practical advantages in this scenario. 
Indeed, they would allow us to run a greater number of planning-execution cycles with minimal human supervision, shorter execution times, and a reduced number of errors in perception or action. 
Moreover, simulations are less susceptible to uncertainty and low-level motion planning failures, which are outside the scope of this work. 
Nonetheless, we preferred to test our architecture with a real robot platform in order to provide a more robust proof-of-concept of the approach.
As a matter of fact, such choice and the related experiences led to the development of the extended scenario, which is described in the next Section.
For these reasons, we provide an example along with a discussion about the limitations of the scenario, and we defer to the extended scenario for a more thorough analysis of our architecture.

Currently, the developed software architecture adopts both off-the-shelf and custom modules.
In particular, we employed ALVAR, a QR tags tracking library typically used in augmented reality applications, to detect the absolute pose of articulated object links using a camera, as well as MoveIt!, which is a \textit{de facto} standard for motion planning and execution in the Robotics community. 
Modules related to ASP have been developed custom.
The architecture has been designed and implemented using the Robot Operating System (ROS, Indigo release) framework, and runs on a machine with an Intel i7-4790 CPU and $16$ GB of RAM.

\begin{figure}[t!]
\centering
$$
\begin{array}{llll}
a_{1}: & \texttt{changeAngle(2,1,90,180,1)} &
\qquad a_{2}: & \texttt{changeAngle(1,0,180,90,2)}\\
a_{3}: & \texttt{changeAngle(3,2,90,180,3)} &
\qquad a_{4}: & \texttt{changeAngle(1,0,270,180,4)}\\
\end{array}
$$
\caption{An excerpt of the answer set returned by Clingo for a problem grounded in the simple scenario. Labels $a_1$ \dots $a_4$ are compact references for ground actions.
\label{fig:valid-asp2}
}
\end{figure}

Tests have been carried out as follows. 
First, the articulated object is located on the table in front of the robot in a \textit{random} configuration that is compatible with the link orientation granularity encoded in the knowledge base, and within an acceptable perceptual error margin.
A video stream is acquired by the camera, QR tag poses are extracted by ALVAR and the \textit{Knowledge Base} module updated accordingly.
It is noteworthy that images are acquired at a $10$ $Hz$ frequency, whereas the knowledge base is updated at a $1$ $Hz$ frequency.
Such a lower frequency update is due to the pre-processing of QR tag poses, which are stabilised using state estimation and filtering techniques.
Initial and goal configurations, as well as information about the length, position and orientation of each link, are therefore represented in terms of the ASP atoms described in Section~\ref{sec:kb}. 

As described above, the \textit{Goal Checker} module continuously control whether the information provided by the perception layer is compatible with the desired articulated object expected configurations.
When this happens, the \textit{Knowledge Base} module is notified, and plan execution proceeds as expected until the goal configuration is reached.
In this case, the \textit{Knowledge Base} module does not produce a new problem instance for the \textit{Action Planning} module, and therefore execution stops. 
Otherwise, if the \textit{Goal Checker} module finds an incoherence in the knowledge representation structure, e.g., the currently perceived articulated object configuration is not compatible with the expected one, then execution is interrupted, and the \textit{Knowledge Base} module updates the problem instance for the \textit{Action Planning} module accordingly.
Such inconsistencies may typically occur in two cases, both generated by rules $c_{12}$, $c_{13}$ and $c_{14}$ shown in Figure~\ref{fig:consistencychecking}. 
As an example, we can consider human-robot cooperation scenarios whereby a human operator and a collaborative robot may interact to jointly manipulate the articulated object \cite{DBLP:journals/ras/CapitanelliMMV18}.
On the one hand, if the robot were loosing grip on a link while manipulating it, or the human operator were rotating a link to an orientation not compatible with the set of represented ones, a consistency exception would rise. 
On the other hand, if the unexpected orientation were compatible with the set of represented ones, plan execution would proceed flawlessly until the robot had to perform an action on that link. 
In such a case, the \textit{Motion Planning} module would expect a certain link orientation different from the current one, which would raise a consistency exception and cause re-planning.

On the basis of the encoding described in Section~\ref{sec:action}, the \textit{Action Planning} module exploits the state-of-the-art ASP solver Clingo~\cite{DBLP:conf/iclp/GebserKKOSW16} to generate a (valid) plan. 
It is important to notice that the video stream from the camera is continuously acquired, and therefore the \textit{Knowledge Base} module updated, and \textit{Consistency Checking} performed.
In order to avoid false positive consistency exceptions, for example during robot manipulation actions, the update of the knowledge base is paused during their execution.

\begin{figure}[t!]
\centering
\subfigure{
\includegraphics[width=0.23\columnwidth]{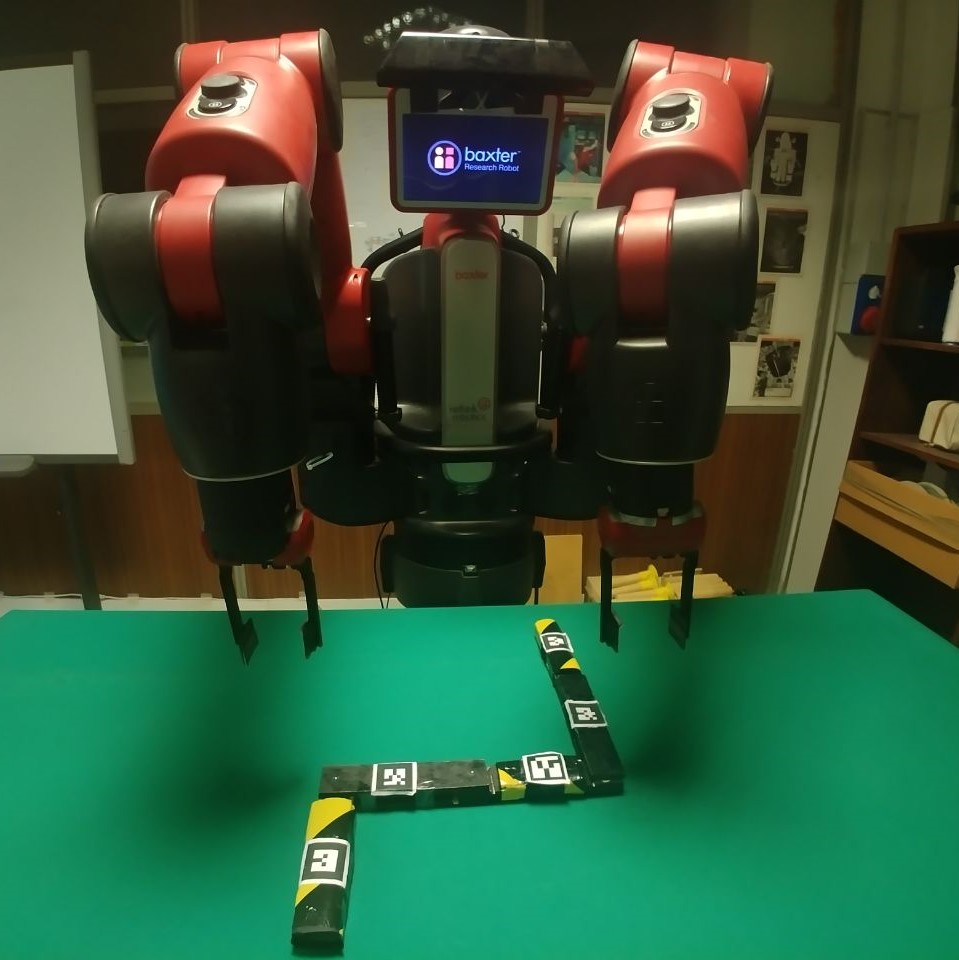}
\label{fig:simple1}}
\subfigure{
\includegraphics[width=0.23\columnwidth]{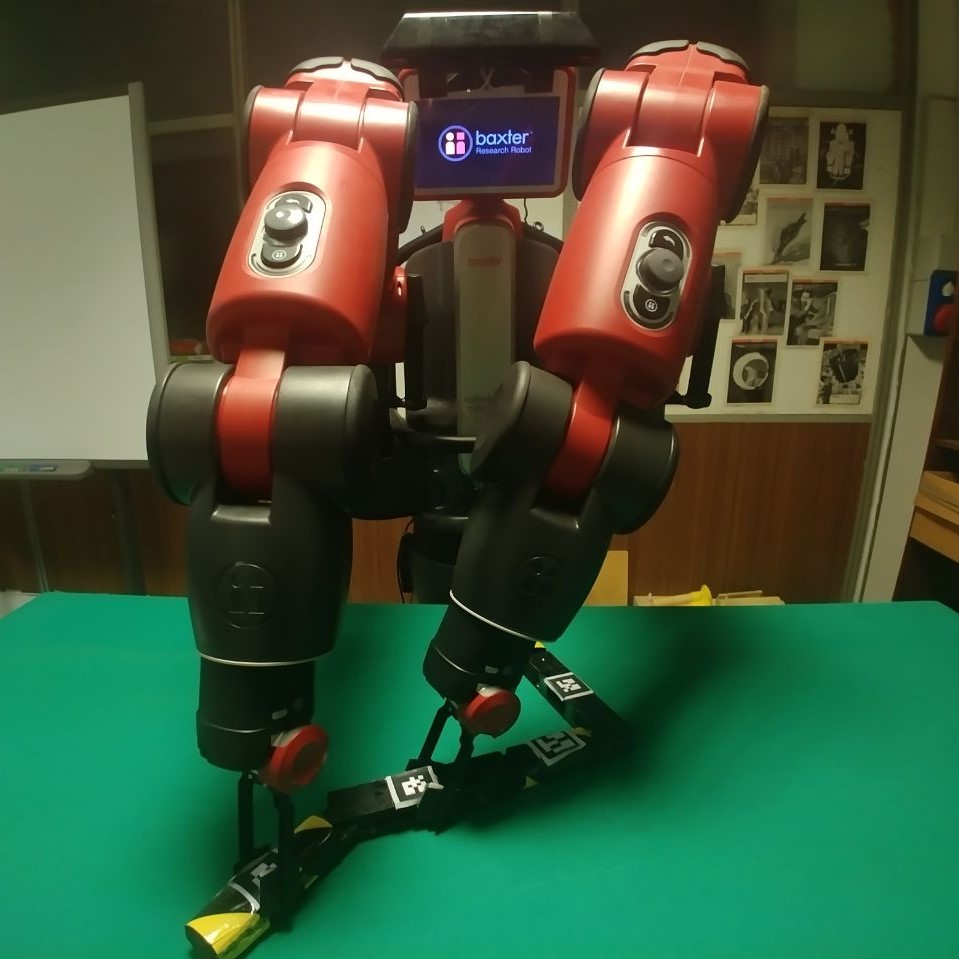}
\label{fig:simple2}}
\subfigure{
\includegraphics[width=0.23\columnwidth]{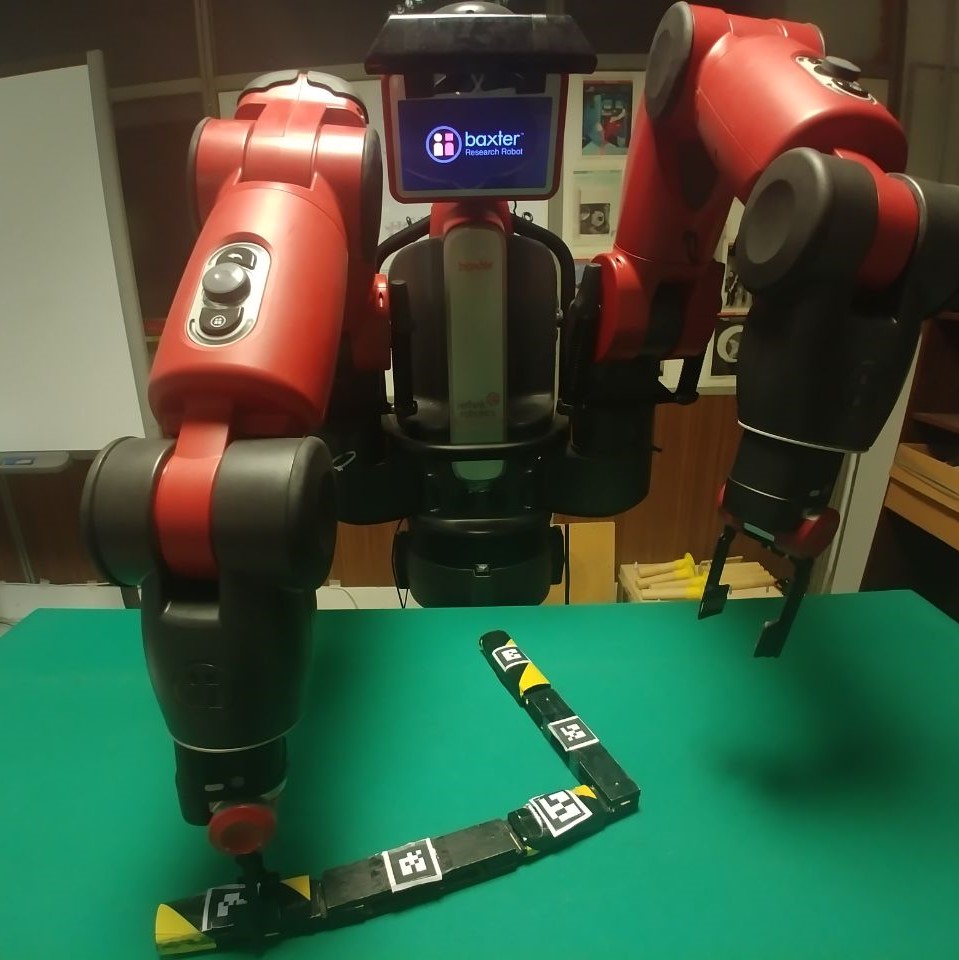}
\label{fig:simple3}}
\subfigure{
\includegraphics[width=0.23\columnwidth]{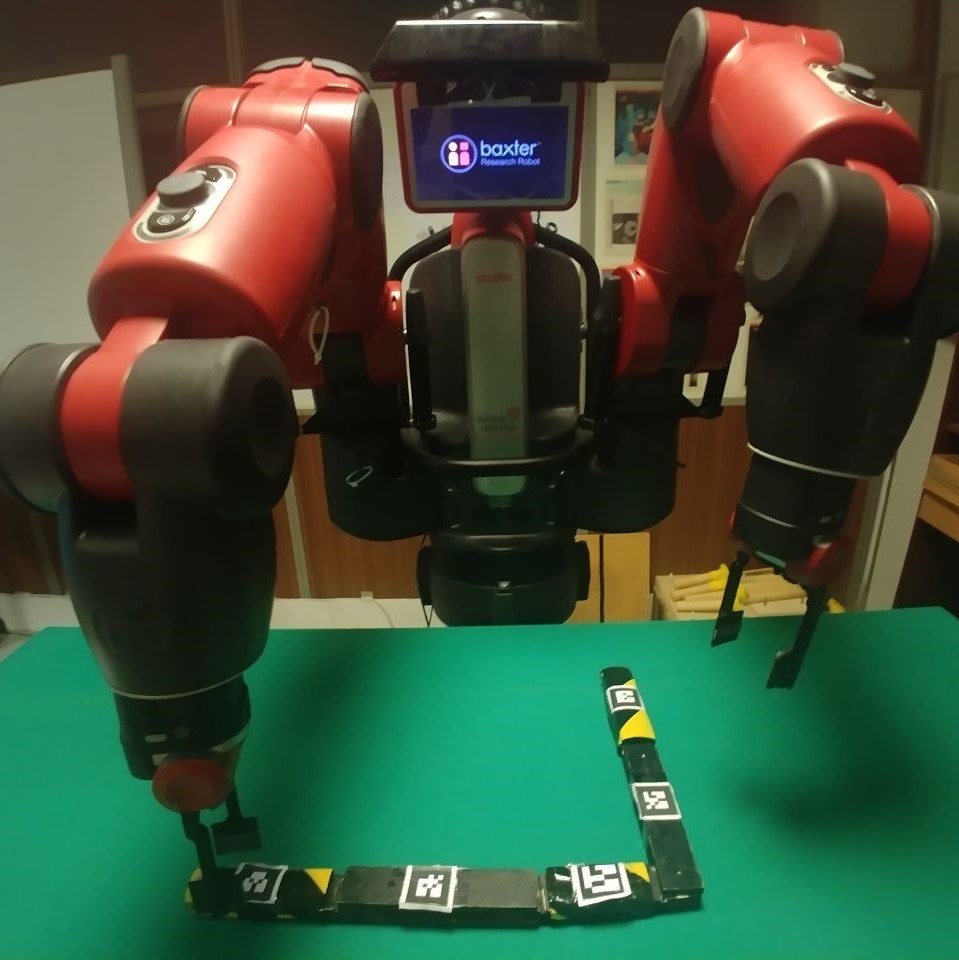}
\label{fig:simple4}}
\subfigure{
\includegraphics[width=0.23\columnwidth]{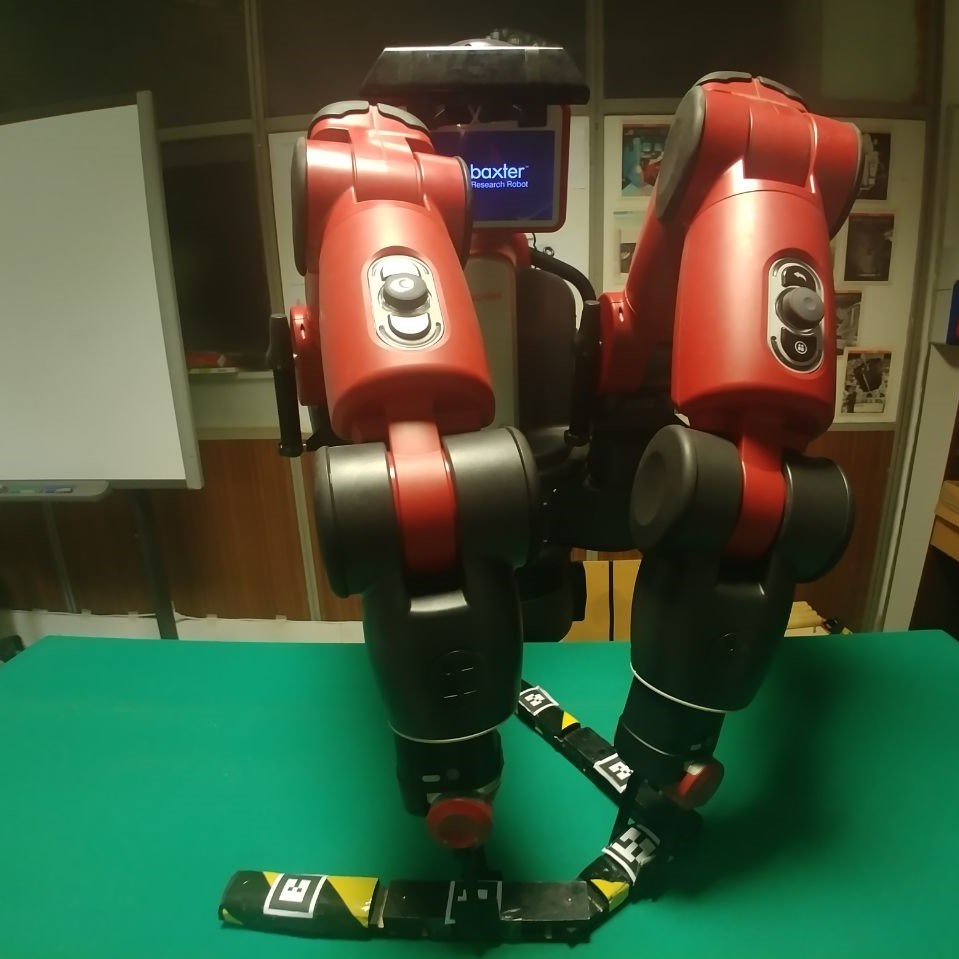}
\label{fig:simple5}}
\subfigure{
\includegraphics[width=0.23\columnwidth]{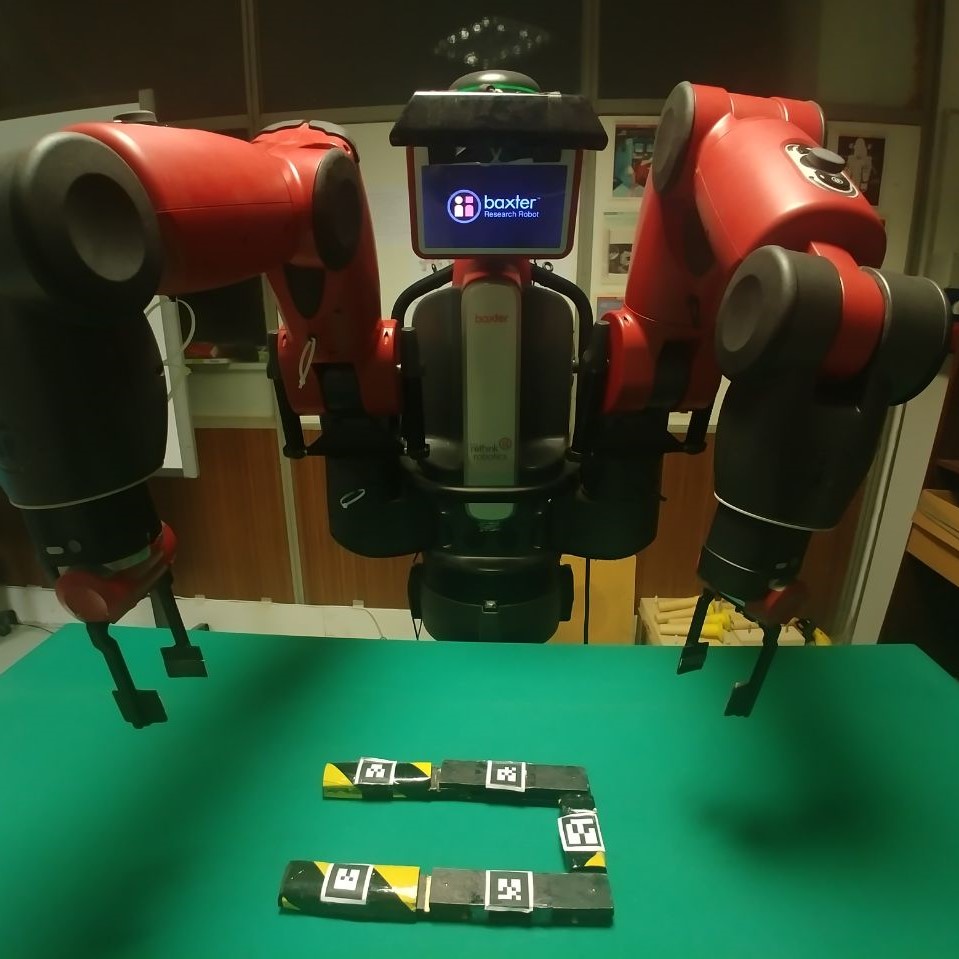}
\label{fig:simple6}}
\subfigure{
\includegraphics[width=0.23\columnwidth]{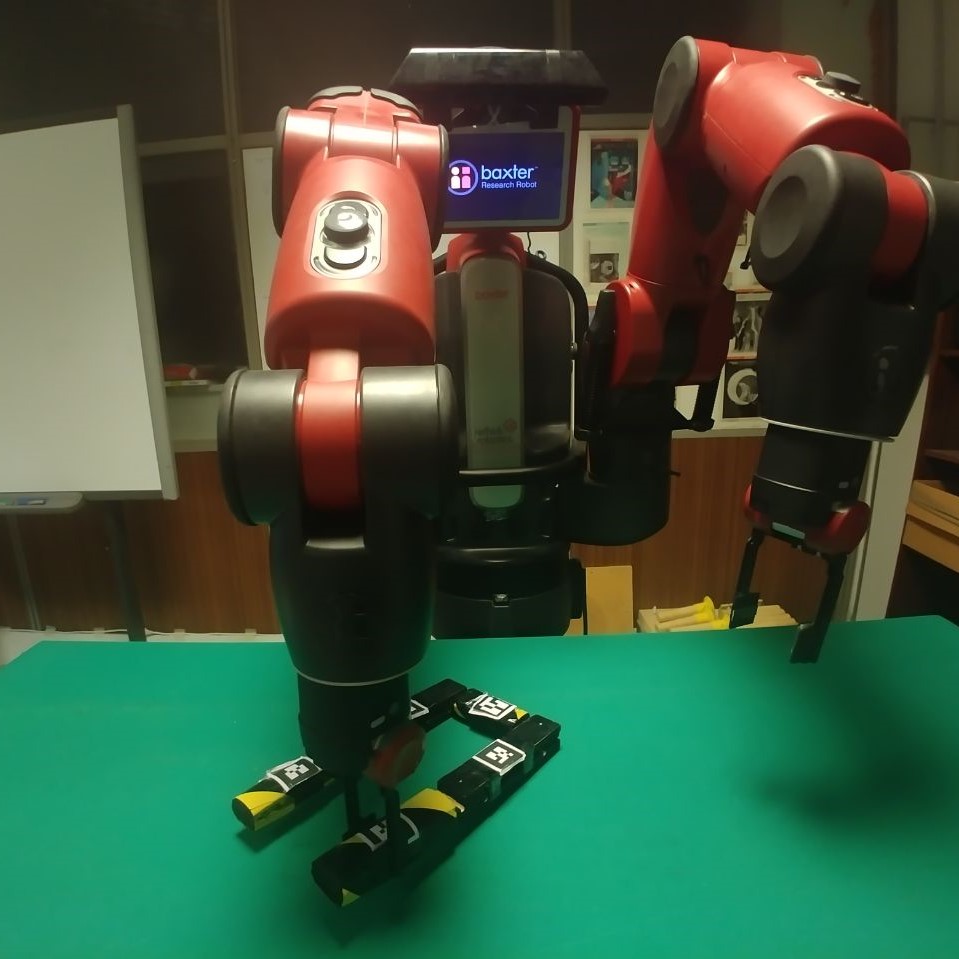}
\label{fig:simple7}}
\subfigure{
\includegraphics[width=0.23\columnwidth]{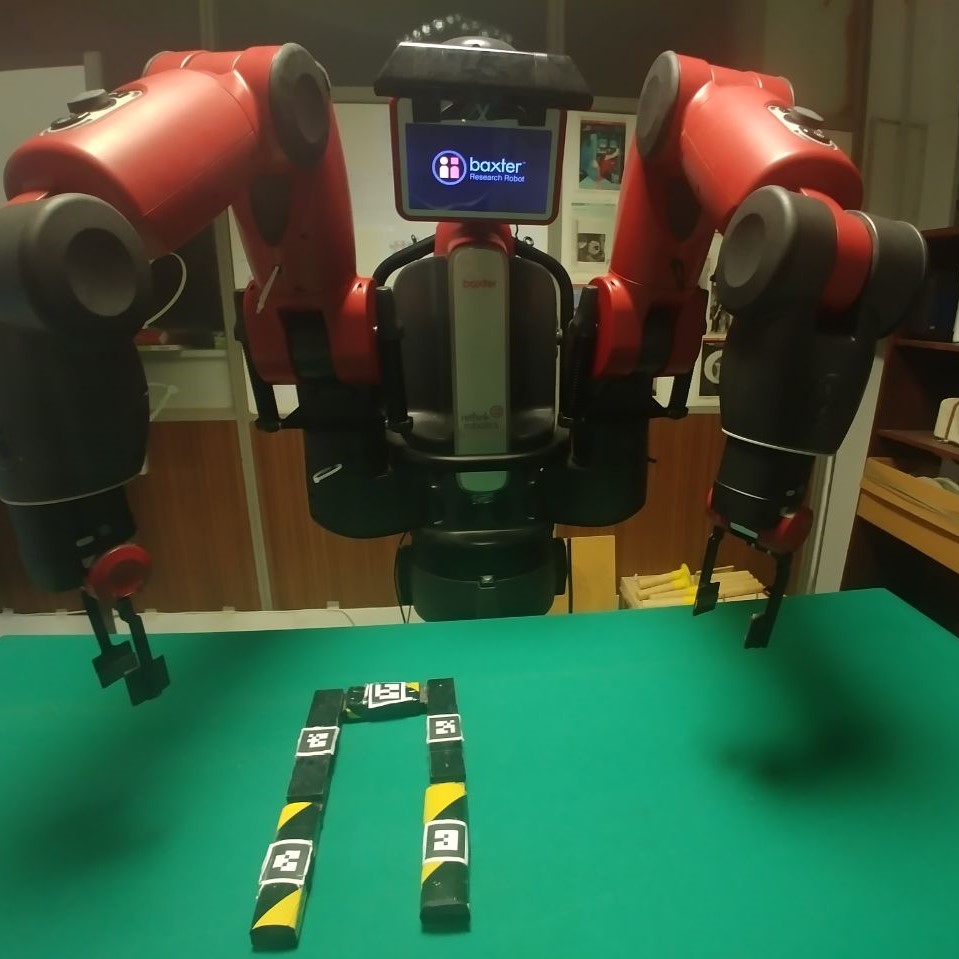}
\label{fig:simple8}}
\caption{Execution of a problem instance in the simple scenario.}
\label{fig:valid-pic}
\end{figure}

%

An example of the whole process is shown in Figure \ref{fig:valid-asp1}, Figure \ref{fig:valid-asp2}, and Figure \ref{fig:valid-pic}.
Figure \ref{fig:valid-asp1} reports an example of ASP-based representation of a problem instance in the simple scenario, whereby the number of joints of the articulated object, their initial poses and the goal configuration to achieve are specified.
Figure \ref{fig:valid-asp2} lists an excerpt of an answer set obtained by Clingo with the encoding presented in Section \ref{sec:action}. 
Each atom of the form \texttt{changeAngle} in the answer set represents an action to perform on a joint, with the meaning detailed in Section \ref{sec:kb}. 
Figure \ref{fig:valid-pic} provides a few snapshots associated with the plan execution process, from top-left to bottom-right. 
In particular, starting from the initial configuration of the articulated object shown in the top-left Figure, the second, fourth, fifth, and seventh Figures represent the execution of actions $a_1$, $a_2$, $a_3$, and $a_4$, respectively, as shown in Figure \ref{fig:valid-asp2}, whereas the third and the sixth represent intermediate configurations. 
Finally, the bottom-right Figure shows the final state corresponding to the articulated object goal configuration.
It is important to note that the fourth Figure displays both $a_3$ execution and its resulting intermediate state since it just corresponds to a rotation of the whole object.

As part of this initial experimental assessment, we performed also a computational  analysis on the performance associated to the \textit{Action Planning} module, by varying the number of links (up to 12), the angle granularity, and by randomly generating initial and final configurations, for a total of 280 instances. 
On the successfully solved problem instances, Clingo took a $1.5s$ processing time on average, and could solve the problems in around $8$ steps on average, with results as low as $0.01$ $s/4$ steps and never above $2.2$ $s/9$ steps, which confirms the applicability of ASP reasoning in this scenario. 
All the plans have been validated with the VAL tool~\cite{DBLP:conf/ictai/HoweyLF04}. 

Albeit the computational performance deteriorates when both the number of links and allowed orientations are increased, results are encouraging considering the limited robot workspace size, and the robot dexterity capabilities for grasping and manipulation, which represent the real drawback of the simple scenario.
As a matter of fact, the majority of run-time issues are related to the impossibility for the robot to reach with both arms the required grasping poses on the articulated object links, due to the fact that rotations \textit{tend} to displace the whole object on the table to a great extent.
It is noteworthy, however, that the proposed ASP-based approach is guaranteed to compute the shortest plan leading to the target object configuration, due to the use of an iterative deepening procedure. 
This is pivotal, as it allows minimising the actual execution time of the plan, which is the most time consuming part of the whole process. 

\section{Extended Scenario and the Updated Encoding}
\label{sec:ext}

Analysing the overall execution performance of our framework in the simple scenario, we noticed that the robot was not always able to perform the required manipulation actions due to the unreachability of the grasping poses associated with the links to manipulate, e.g., for links displaced outside the robot manipulation space.

These considerations led us to the design of a new, extended scenario.
The scenario does not modify any physical characteristics with respect to the setup introduced in Section \ref{label:srs}, but represents the setup and its constraints with a higher accuracy with the aim of enforcing actions feasibility. 
In particular, the extended scenario builds up on a model with an improved trade-off between accurately modelling the physics of the specific scenario, and having a general model that can be easily reused in similar settings. 
To this aim, the most sensible aspect of the scenario at hand is related to links manipulation, so in the new encoding it is modelled with a greater level of detail.
Furthermore, there are different ways whereby a robot can centre a link to manipulate with respect to its manipulation space. 
For this reason, the newly introduced encoding provides a more general way for modelling this specific aspect. 
On top of these considerations, there is still the need for the model to be operational, so that ASP solvers can find solutions in a reasonable amount of time. 

In this Section we present, in two separate sub-sections, the extended scenario we have designed, and the related, improved encoding we have defined in order to deal with such scenario in our robot control architecture. 

\subsection{Extended Scenario}

Herewith we briefly describe the main differences with respect to the simple scenario.
These can be summarised as follows.

\begin{itemize}
\item
\textit{Robot grippers are explicitly modelled}.
Each robot gripper should be considered as a resource that can be either  \textit{occupied}, e.g., keeping a link firmly or rotating a link, or \textit{free}. 
With this modification, it is possible to explicitly represent which robot gripper can manipulate a given link. 
\item
In the simple scenario, it may happen that a link could not be reached or grasped, or it may be placed to an area of the robot's workspace where manipulation could be difficult or even impossible because of the robot kinematics configuration.
This situation may happen for reasons related to the articulated object's configuration while being manipulated, or the specific sequence of manipulation actions required by the plan.
In the extended scenario, each time a manipulation action is carried out on a given link, \textit{it is assured that the target link is centred in the robot workspace}. 
If this is not the case, the link is first moved towards the central part of the table.
This maximises the likelihood of a relevant link to be successfully reached and manipulated by the robot.
\item
\textit{Grasping and release actions by the two grippers are explicitly modelled}.
In the scenario described in Section \ref{label:srs}, these two manipulation actions were not modelled, although they were assumed to be properly carried out during the execution phase, as part of each action execution.
By considering an explicit modelling, we can represent grippers occupancy, and we can better characterise the semantics associated with each action, since now grasping, manipulation, and release are distinct.
\end{itemize}

It is noteworthy that the above mentioned features allow for a number of improvements. 
We can envisage two main advantages.
On the one hand, this encoding is expected to better manage the explicitly modelled robot resources, i.e., the grippers.
On the other hand, manipulation actions are now characterised by a more precise semantics, which does not make any implicit assumption about actual robot behaviour. In a nutshell, the encoding provides a more accurate description of the domain, that supports a less abstract planning process, and can easily generalise to different conditions (e.g, by modelling robot resources, you can also deal with cases where some of those resources are unavailable).

Furthermore, in the new encoding we provide a specific action for displacing links to the centre of the robot manipulation space, which enforces the reasoner to take positioning into account, but does not explicitly mention the sequence of movements to be done for this operation. 
Such choice has been taken for two main reasons: 
$(i)$ different robots may employ different strategies to displace the object, e,g., using a combination of manipulation and pushing behaviours, so we aimed at providing a model that can be easily re-used with other platforms, and 
$(ii)$ the way a link is moved to the centre of the workspace requires a somewhat lower degree of accuracy when compared to a manipulation action.

\subsection{ASP Encoding for the Extended Scenario}
\label{sub_asp_modules}

In this Section we describe ASP-related modifications to the \textit{Knowledge Base} module and the \textit{Action Planning} module to implement the extended scenario.

\paragraph{\it ASP encoding for the Knowledge Base module.}
\label{subsec:kb}

In order to have a working architecture in the extended scenario we have to update the \textit{Knowledge Base} module, which has been extended to include more atoms.
Atoms \texttt{gripper(1)} and \texttt{gripper(2)} represent the two robot grippers.
Atoms \texttt{free(1,0)} and \texttt{free(2,0)} represent the fact that at time step $0$ the two robot grippers are free and ready to grasp.
Atom \texttt{in\_centre(J,0)} represents the fact that the joint \texttt{J} is at the centre of the workspace at time step $0$.
Moreover, this last change led us to the necessity to add to the knowledge base both links and joints, because the robot must act on the links but we want that a certain joint is in the middle of the workspace. 
For this reason, the atoms of the form \texttt{isLinked(L1,L2)}, where \texttt{L1} and \texttt{L2} are two links, became \texttt{connected(J,L)}, where \texttt{J} is a joint and \texttt{L} a link. 
We also use facts over atoms of the form \texttt{link(L)} to represent the links of the articulated object.
Finally, we mention that atoms of the form \texttt{hasAngle(J,A,T)}, where the first term \texttt{J} represents a joint, are replaced by atoms of the form \texttt{hasAngle(L,A,T)}, where the first term \texttt{L} is a link.

\paragraph{\it ASP encoding for the Consistency Checking module.}
The changes to the \textit{Knowledge Base} module require also to update the \textit{Consistency Checking} module, in order to keep consistency with the other modules of the architecture. 
For this reason, we add a check on each new added atom and modified rules that were not needed for the new knowledge base. Such checks are similar to the ones presented in Figure~\ref{fig:consistencychecking} and reported as follows:
$$
\begin{array}{ll}
c_{3a'}\ & \texttt{\footnotesize{:- hasAngle(L,A,T), not link(L).}}\\
c_{15}\ & \texttt{\footnotesize{:- connected(J,L), not joint(J).}}\\
c_{16}\ & \texttt{\footnotesize{:- connected(J,L), not link(L).}}\\
c_{17}\ & \texttt{\footnotesize{:- in\_centre(J,T), not joint(J).}}\\
c_{18}\ & \texttt{\footnotesize{:- in\_centre(J,T), not time(T).}}\\
c_{19}\ & \texttt{\footnotesize{:- in\_hand(L,T), not link(L).}}\\
c_{20}\ & \texttt{\footnotesize{:- in\_hand(L,T), not time(T).}}\\
c_{21}\ & \texttt{\footnotesize{:- grasped(G,L,T), not gripper(G).}}\\
c_{22}\ & \texttt{\footnotesize{:- grasped(G,L,T), not link(L).}}\\
c_{23}\ & \texttt{\footnotesize{:- grasped(G,L,T), not time(T).}}\\
c_{24}\ & \texttt{\footnotesize{:- free(G,T), not gripper(G).}}\\
c_{25}\ & \texttt{\footnotesize{:- free(G,T), not time(T).}}\\
\end{array}
$$
Note that $c_{1a}$, $c_{1b}$, and $c_{2}$ are removed since \texttt{isLinked(L1,L2)} are not part of the Knowledge Base, whereas $c_{3}$ is replaced with $c_{3a'}$. 

\paragraph{\it ASP encoding for the Action Planning Module.}
\label{subsec:SAES}

\begin{figure}[t!]
$$
\begin{array}{ll}
r_{5'}\ & \texttt{\{changeAngle(L1,L2,J1,A1,A2,G1,G2,T)\} :- link(L1), link(L2),}\\
& \qquad \texttt{joint(J1), angles(A1), angles(A2), gripper(G1), gripper(G2),}\\
& \qquad \texttt{time(T), in\_centre(J1,T), grasped(G1,L1,T), grasped(G2,L2,T),}\\
& \qquad \texttt{in\_hand(L1,T), in\_hand(L2,T), not free(G1,T),}\\
& \qquad \texttt{not free(G2,T), connected(J1,L1), connected(J1,L2),}\\
& \qquad \texttt{hasAngle(L1,A2,T), L1<>L2, G1<>G2}
\end{array}
$$
\caption{Rule $r_{5}$ with its preconditions updated.}
\label{fig:rule5}
\end{figure}

\begin{figure}[t!]
$$
\begin{array}{ll}
r_{16}\ & \texttt{\{move\_link\_to\_central(L1,J1,G2,T)\}:-link(L1), joint(J1),}\\
 & \qquad \texttt{gripper(G2), time(T), connected(J1,L1), free(G2,T), }\\
 & \qquad \texttt{not in\_hand(L1,T), not in\_centre(J1,T).}\\
r_{17}\ & \texttt{in\_centre(J1,T+1):-move\_link\_to\_central(\_,J1,\_,T), T<timemax+1.}\\
r_{18}\ & \texttt{\{take\_links\_to\_move(L1,L2,J1,G1,G2,T)\}:-link(L1), link(L2),}\\
& \qquad \texttt{joint(J1), gripper(G1), gripper(G2), in\_centre(J1,T)}\\
& \qquad \texttt{free(G1,T), free(G2,T), not in\_hand(L1,T), not in\_hand(L2,T),}\\
& \qquad \texttt{connected(J1,L1), connected(J1,L2), L1<>L2, G1<>G2.}\\
r_{19}\ & \texttt{in\_hand(L1,T+1):-take\_links\_to\_move(L1,\_,\_,\_,\_,T), T<timemax+1.}\\
r_{20}\ & \texttt{in\_hand(L2,T+1):-take\_links\_to\_move(\_,L2,\_,\_,\_,T), T<timemax+1.}\\
r_{21}\ & \texttt{grasped(G1,L1,T+1):-take\_links\_to\_move(L1,\_,\_,G1,\_,T), T<timemax+1.}\\
r_{22}\ & \texttt{grasped(G2,L2,T+1):-take\_links\_to\_move(\_,L2,\_,\_,G2,T), T<timemax+1.}\\
r_{23}\ & \texttt{\{release\_links(L1,L2,J1,G1,G2,T)\}:-link(L1), link(L2), joint(J1),} \\
& \qquad \texttt{gripper(G1), gripper(G2), time(T), grasped(G2,L2,T),}\\
& \qquad \texttt{grasped(G1,L1,T), in\_hand(L1,T), in\_hand(L2,T),}\\
& \qquad \texttt{not free(G1,T), not free(G2,T), connected(J1,L1),}\\
& \qquad \texttt{connected(J1,L2), L1<>L2,G1<>G2.}\\
r_{24}\ & \texttt{free(G2,T+1):-release\_links(\_,\_,\_,\_,G2,T),T<timemax+1.}\\
r_{25}\ & \texttt{free(G1,T+1):-release\_links(\_,\_,\_,G1,\_,T),T<timemax+1.}\\
r_{26}\ & \texttt{action(T,move\_link\_to\_central(L1,J1,G2,T)) :-} \\
& \qquad \texttt{move\_link\_to\_central(L1,J1,G2,T).}\\
r_{27}\ & \texttt{action(T,take\_links\_to\_move(L1,L2,J,G1,G2,T)) :-}\\
& \qquad \texttt{take\_links\_to\_move(L1,L2,J,G1,G2,T).}\\
r_{28}\ & \texttt{action(T,changeAngle(L1,L2,J,A1,A2,G1,G2,T)) :-} \\
& \qquad \texttt{changeAngle(L1,L2,J,A1,A2,G1,G2,T).}\\
r_{29}\ & \texttt{action(T,release\_links(L1,L2,J1,G1,G2,T)) :-} \\
& \qquad \texttt{release\_links(L1,L2,J1,G1,G2,T).}\\
r_{30}\ & \texttt{:- time(T), \#count\{Z : action(T,Z)\} != 1.}\\
r_{31}\ & \texttt{in\_hand(L,T+1):-in\_hand(L,T), not release\_links(L,\_,\_,\_,\_,T),}\\
& \qquad \texttt{not release\_links(\_,L,\_,\_,\_,T), T<timemax+1.}\\
r_{32}\ & \texttt{free(G,T+1):-free(G,T), not take\_links\_to\_move(\_,\_,\_,\_,G,T),}\\
& \qquad \texttt{not take\_links\_to\_move(\_,\_,\_,G,\_,T), T<timemax+1.}\\
r_{33}\ & \texttt{grasped(G,L,T+1):-grasped(G,L,T), T < timemax+1}\\
& \qquad \texttt{not release\_links(L,\_,\_,G,\_,T), not release\_links(\_,L,\_,\_,G,T).}\\
\end{array}
$$
\caption{Simple Action Extended Scenario (SAES) encoding.}
\label{fig:extended_s}
\end{figure}

Differently from the encoding we designed for the simple scenario (Figure~\ref{fig:standard_f}), and which from now on we refer to as \textit{Simple Action Scenario} (SAS), in the encoding for the extended scenario, referred to as \textit{Simple Action Extended Scenario} (SAES) from now on, we added the possibility of performing manipulation actions in both directions, i.e., it is now possible to rotate a link with respect to any of its neighbouring links in the articulated object chain.
This is accomplished by slightly modifying the encoding to allow for both forward and backward propagation. 
In particular, we removed the constraint $J1 > J2$ in rule $r_{5}$ and we changed its body. 
Such updates are necessary to comply with the new model: since two links must be grasped we had to consider the presence of the grippers and of the links. Moreover, those links must be at the centre of the robot manipulation space and, therefore, it is necessary to ensure that the rule is selected only if the joint between them is located at the centre. 
For the sake of readability we will, from now on, refer to the modified version of rule $r_{5}$ as rule $r_{5'}$, as shown in Figure \ref{fig:rule5}.
It is important to emphasise here that, aside from such changes, the idea behind the SAS encoding is maintained: the general structure and the strategy employed by the SAES encoding to deal with manipulation actions related to the articulated object configuration changes are the same as in Figure~\ref{fig:standard_f}.

Figure~\ref{fig:extended_s} reports the additional rules needed to deal with the extended scenario.
Rules $r_{16}$, $r_{18}$, and $r_{23}$ are related to the selection of possible actions in this model: 
$r_{16}$ locates the joint that has to be moved to the centre of the manipulation space, $r_{18}$ selects an atom of the form \texttt{take\_links\_to\_move(L1,L2,J,G1,G2,T)}, where \texttt{L1} and \texttt{L2} are the links that have to be grasped, while $r_{23}$ selects an atom of the form \texttt{release\_links(L1,L2,J,G1,G2,T)} to release the links the robot is acting upon, respectively.
Rule $r_{17}$ is used to signal which link is located at the centre of the space, as effects of \texttt{move\_link\_to\_central(L1,J1,G2,T)}. 
To this end, we use atoms of the form \texttt{in\_centre(J,T)}, which represent the fact that a joint \texttt{J} is at the centre at time step \texttt{T}.
Rules $r_{19}$ and $r_{20}$ are used to identify grasped links.
In particular, they use atoms of the form \texttt{in\_hand(L,T)}, representing the fact that a link \texttt{L} has been grasped by one robot gripper at time step \texttt{T}.
In a similar way, $r_{21}$ and $r_{22}$ are used to identify which gripper is occupied or is free.
To this aim, atoms of the form \texttt{grasped(G,L,T)} are used, and they represent the fact that a link \texttt{L} is grasped at the time step \texttt{T} using gripper \texttt{G}. 
Rules $r_{24}$ and $r_{25}$ are used to notify that robot grippers are free and they can be used again.
We used atoms of the form \texttt{free(G,T)}, which model the fact that a gripper \texttt{G} is free at time step \texttt{T}.
Furthermore, rules from $r_{26}$ to $r_{30}$ ensure that only one action (among the ones represented by rules $r_{16}$, $r_{18}$, $r_{23}$, and $r_{5'}$) is selected for each time step.
Atoms of the form \texttt{action(T,A)} are used to ensure that action \texttt{A} is executed at time step \texttt{T}.
Finally, the rules from $r_{31}$ to $r_{33}$ are used to propagate the information that has not been changed in the current time step to the next one. 
It is noteworthy that grasping for \textit{manipulating} links or grasping for \textit{displacing} a link to the centre of the manipulation space are very different actions.
Links must be grasped in different ways as the robot has to move the arms differently.
For this reason, our encoding requires that after the link to be moved has been put in the centre of the workspace (rule $r_{16}$), appropriate grasp actions have to be performed before actual manipulation.
Overall, the SAES encoding is composed by:
\begin{itemize}
    \item the rules from $r_{1}$ to $r_{4}$ and from $r_{6}$ to $r_{15}$ of \textit{SAS} (shown in Figure~\ref{fig:standard_f}), 
    \item the rule $r_{5'}$ shown in Figure~\ref{fig:rule5}, and
    \item the rules from $r_{16}$ to $r_{33}$ shown in Figure \ref{fig:extended_s}.
\end{itemize}
The rule involving the goal is already present in the \textit{SAS} encoding, i.e., $r_{15}$. 

\section{Macro Actions and their Usage in the Extended Scenario}
\label{sec:macro}

This Section shows how the concept of macro actions, henceforth possibly referred to simply as \textit{macros}, originally defined in the Automated Planning community, can be fruitfully employed in the context of the extended scenario. 
The Section describes the general concept of what a macro is, how macros can be represented in ASP, and a description about the macros we have employed and their encoding, respectively.

\subsection{Macros in Automated Planning}
\label{subsec:map}

Macros encapsulate sequences of elementary planning operators.
Advantageously, they can be encoded in the same form as planning operators, and therefore they can be added into a domain model seamlessly and can be exploited in a \textit{solver-independent} way. 
Macros can be seen as \textit{short-cuts} in the state space of a planning problem.
They can reduce the number of steps needed to reach the goal state, however at the cost of an increased branching factor. 
It is noteworthy that macros can provide a mean for encoding additional, practical and common sense knowledge about a specific domain.
A domain expert can suggest actions that are usually executed in sequence to be encoded as a macro and added to the domain model. 
On the one hand, if macros are added to the original model, and the related elementary operators are not removed, the extended model is able to cope with cases where encapsulated operators are \textit{usually} executed in sequence, but not always. 
On the other hand, removing the original operators can lead to more significant performance improvements. 
It is noteworthy that macros could be considered as a sort of \textit{extra-logic} trick to force a planning engine towards certain trajectories in the planning search space.
As such, one may argue whether there could be a better option to implement that trick, e.g., in the action execution phase of our architecture.
While it may be certainly possible to achieve that, for instance by performing a post-processing step on a plan made up of elementary actions, that would make the architecture less flexible and general-purpose.
Instead, macro encodings can be seamlessly integrated with planning operators, and therefore the engineering effort to add, remove, or modify them is minimum.

For the sake of the discussion carried out in the Section, let us provide a more formal description of planning operators, and of macros in automated planning.
We say that an operator
\begin{equation}
o = (\MathWord{name}(o), \MathWord{pre}(o), \MathWord{del}(o), \MathWord{add}(o))    
\end{equation}
is a \textit{planning operator}, where 
$\MathWord{name}(o) = \MathWord{op\_name}(x_1, \dots, x_k)$, with $\MathWord{op\_name}$ is a unique operator name and $x_1,\dots x_k$ are $k$ variable symbols, i.e., arguments, appearing in the operator, and
$\MathWord{pre}(o),\MathWord{del}(o)$ and $\MathWord{add}(o)$ are sets of (non) grounded predicates with variables taken only from $x_1, \dots, x_k$ representing operator precondition, delete, and add effects, respectively. 
Actions are \textit{grounded instances} of planning operators. 
An action $a$ is \textit{applicable} in a state $s$ if and only if $\MathWord{pre}(a) \subseteq s$. 
The application of $a$ in $s$, if possible, results in a state $(s\setminus\MathWord{del}(a))\cup \MathWord{add}(a))$; we call $(\MathWord{del}(a))\cup \MathWord{add}(a))$ \texttt{effects}.
%
%
Therefore, a macro $o_{i,j}$ is constructed by orderly assembling planning operators $o_i$ and $o_j$ as follows. 
Let $\Phi$ and $\Psi$ be mappings between variable symbols, because we may need to appropriately rename variable symbols of $o_i$ and $o_j$ to construct $o_{i,j}$, the macro is constructed as follows: 
\begin{itemize}
\item
$\MathWord{pre}(o_{i,j})=\MathWord{pre}(\Phi(o_i))\cup (\MathWord{pre}(\Psi(o_j))\setminus \MathWord{add}(\Phi(o_i)))$,
\item
$\MathWord{del}(o_{i,j})=(\MathWord{del}(\Phi(o_i))\setminus \MathWord{add}(\Psi(o_j)))\cup \MathWord{del}(\Psi(o_j))$,
\item
$\MathWord{add}(o_{i,j})=(\MathWord{add}(\Phi(o_i))\setminus \MathWord{del}(\Psi(o_j)))\cup \MathWord{add}(\Psi(o_j))$.
\end{itemize}
Longer macros, i.e., macros encapsulating longer sequences of original planning operators, can be constructed iteratively using the same approach.

For a macro to be \textit{sound}, no instance of $\Phi(o_{i})$ can delete an atom required by a corresponding instance of $\Psi(o_{j})$, 
otherwise they cannot be applied consecutively, whereas it is obvious that if a predicate deleted by $\Phi(o_i)$ (and not added back) is the same (in terms of both name and variable symbols) as a predicate in the precondition of $\Psi(o_j)$, then the macro $o_{i,j}$ is unsound. 

\subsection{Representing Macros in ASP}
\label{subsec:masp}
In this Section we describe how macros are encoded in ASP. In order to ease the descriptions, we emphasize the relation between our SAES encoding and common concepts in planning.
In particular, atoms appearing in the head of rules $r_{16}$, $r_{18}$, and $r_{23}$ represent the \textit{actions}, whereas their bodies represent \textit{preconditions} that must be satisfied in order to select that action.
Rule $r_{17}$ represents the \textit{effect} of the action described in $r_{16}$; rules $r_{19}$-$r_{22}$ and $r_{32}$ represent the effects of the action in rule $r_{18}$; and rules $r_{24}$-$r_{25}$, $r_{31}$, and $r_{32}$ represent the effects of the action in rule $r_{23}$.
\textit{Fluents} are all the atoms appearing in the encoding that contains the variable representing the time, in our case usually referred to as \texttt{T}.

In ASP, a macro is encoded by a single \textit{choice} rule, which implicitly represents multiple actions, and by several normal rules needed to model their effects.
The \textit{head} of the choice rule contains a single fresh atom, whose variables are all the one appearing in the body of the choice rule, whereas its \textit{body} is built as described in the following paragraph.
Given a rule $r_i$ representing an action, $\MathWord{pre}(r_i)$ denotes the rule body.
Intuitively, it represents the conditions that must hold in order to activate the action represented by the rule.
Moreover, $\MathWord{del}(r_i)$ (respectively, $\MathWord{add}(r_i)$) represent all the atoms which are set as false (respectively, true) whenever the conditions denoted by $\MathWord{pre}(r_i)$ hold.
Therefore, a macro $r_{i,j}$ is constructed by assembling the rules representing single actions and by generating $\MathWord{pre}(r_{i,j})$, $\MathWord{del}(r_{i,j})$, and $\MathWord{add}(r_{i,j})$, as follows:
\begin{itemize}
\item
$\MathWord{pre}(r_{i,j})=\MathWord{pre}(r_i)\cup (\MathWord{pre}(r_j)\setminus \MathWord{add}(r_i))$,
\item 
$\MathWord{del}(r_{i,j})=(\MathWord{del}(r_i)\setminus \MathWord{add}(r_j))\cup \MathWord{del}(r_j)$,
\item $\MathWord{add}(r_{i,j})=(\MathWord{add}(r_i)\setminus \MathWord{del}(r_j))\cup \MathWord{add}(r_j)$.
\end{itemize}
where $r_i$ and $r_j$ are two distinct rules. 
As a consequence, for a macro $r_{i,j}$, the body of the choice rule is represented by $\MathWord{pre}(r_{i,j})$. 

\begin{Example}
\label{ex:macro}
As an example, let us consider the two rules $r_{16}$ and $r_{18}$, which correspond to the two actions \texttt{move\_link\_to\_central} and \texttt{take\_links\_to\_move}.
The aim is to create the macro \texttt{linkToCentral\_take} combining such two actions.
Therefore, $\mathit{pre}(r_{16})$ and $\mathit{pre}(r_{18})$ correspond to the bodies of rules $r_{16}$ and $r_{18}$, i.e., they are defined as follows:
$$
\small
\begin{array}{ll}
\mathit{pre}(r_{16}) =& \{\texttt{link(L1)}, \texttt{joint(J1)}, \texttt{gripper(G2)},  \texttt{time(T)}, \texttt{free(G2,T)}, \\
& \texttt{connected(J1,L1)}, \texttt{not in\_hand(L1,T)}, \texttt{not in\_centre(J1,T)}\}\\
\mathit{pre}(r_{18}) =& \{\texttt{link(L1)}, \texttt{link(L2)}, \texttt{joint(J1)}, \texttt{gripper(G1)}, \texttt{gripper(G2)},\\
& \texttt{in\_centre(J1,T)}, \texttt{free(G1,T)}, \texttt{free(G2,T)}, \texttt{connected(J1,L2)}, \\
& \texttt{not in\_hand(L1,T)}, \texttt{not in\_hand(L2,T)}, \texttt{connected(J1,L1)},\\
& \texttt{L1<>L2}, \texttt{G1<>G2}\}.\\
\end{array}
$$
Afterwards, $\mathit{del}(r_{16})$ and $\mathit{del}(r_{18})$ contain all the atoms which are set to false after the corresponding actions have been selected. 
In our example, no atoms are set to false because action \texttt{move\_link\_to\_central} is selected, therefore $\mathit{del}(r_{16}) = \emptyset$.
Instead, \texttt{free(G,T)} is set to false because \texttt{take\_links\_to\_move} is selected from rule $r_{32}$, i.e., since only one action can be selected at each time step.
Therefore, $\mathit{del}(r_{18})=\{\texttt{free(G,T)}\}$.
Finally, $\mathit{add}(r_{16})$ and $\mathit{add}(r_{18})$ correspond to atoms which are set to true after the corresponding actions have been selected.
In our example, action \texttt{move\_link\_to\_central} leads to atom \texttt{in\_centre(J,T)} holding true (from rule $r_{17}$), whereas \texttt{take\_links\_to\_move} leads to atom \texttt{in\_hand(L1,T)}, \texttt{in\_hand(L2,T)} (from rules $r_{19}$ and $r_{20}$), \texttt{grasped(G1,L1,T)} and \texttt{grasped(G2,L2,T)} (from rules $r_{21}$ and $r_{22}$) holding true.
Therefore:
$$
\begin{array}{ll}
\mathit{add}(r_{16}) =& \{\texttt{in\_centre(J1,T)}\}\\
\mathit{add}(r_{18}) =& \{\texttt{in\_hand(L1,T)}, \texttt{in\_hand(L2,T)},\\
& \texttt{grasped(G1,L1,T)}, \texttt{grasped(G2,L2,T)}\}.\\
\end{array}
$$
Now we are ready to define $\mathit{pre}(r_{16,18})$, $\mathit{add}(r_{16,18})$, and $\mathit{del}(r_{16,18})$, which are as follows:
$$
\begin{array}{ll}
\mathit{pre}(r_{16,18}) =& \{\texttt{link(L1)},  \texttt{link(L2)}, \texttt{joint(J1)}, 
 \texttt{gripper(G1)}, \texttt{time(T)},\\
& \texttt{gripper(G2)}, \texttt{not in\_centre(J1,T)}, \texttt{free(G1,T)}, \\
& \texttt{free(G2,T)},
 \texttt{not in\_hand(L1,T)}, \texttt{not in\_hand(L2,T)}\\
& \texttt{connected(J1,L1)}, \texttt{connected(J1,L2)}, \texttt{L1<>L2}, \texttt{G1<>G2}\}\\
\mathit{del}(r_{16,18}) =& \{\texttt{free(G,T)}\}\\
\mathit{add}(r_{16,18}) =& \{\texttt{in\_centre(J1,T)}, \texttt{in\_hand(L1,T)}, \texttt{in\_hand(L2,T)},\\
& \texttt{grasped(G1,L1,T)}, \texttt{grasped(G2,L2,T)}\}.\\
\end{array}
$$
Therefore the choice rule representing the macro $r_{16,18}$ is as follows:
$$
\begin{array}{ll}
& \texttt{\{linkToCentral\_take(L1,L2,J1,G1,G2,T)\} :-} \texttt{link(L1)},  \texttt{link(L2)},\\ 
& \qquad \texttt{gripper(G1)}, \texttt{gripper(G2)}, \texttt{time(T)}, \texttt{free(G1,T)}, \texttt{free(G2,T)},\\
& \qquad \texttt{not in\_centre(J1,T)}, \texttt{not in\_hand(L1,T)}, \texttt{not in\_hand(L2,T)},\\
& \qquad \texttt{L1<>L2}, \texttt{G1<>G2},\texttt{joint(J1)},\texttt{connected(J1,L1)}, \texttt{connected(J1,L2)}.\\
\end{array}
$$
\end{Example}

\subsection{Macros for the Extended Scenario and their ASP Encoding}
\label{subsec:MAES}

While evaluating robot plans related to problem instances in the extended scenario, we noticed that some of the actions could be merged. 
In fact, we observed typical sequences of actions corresponding to the manipulation of links. 
Leveraging on this observation, we modified the SAES encoding to include such macros. From now on we will refer to the new encoding as \textit{Macro Action Extended Scenario} (MAES).
This Section describes the macros we have defined, and then how they have been encoded in ASP, following the example given in the previous Subsection.
The complete MAES encoding can be found in Appendix A.  

\paragraph{\it Macros for the Extended Scenario.}
The following macros have been considered: 
\begin{itemize}
\item 
\texttt{linkToCentral\_take} is shown in Example~\ref{ex:macro}, and it is the composition of the action \texttt{move\_link\_to\_central}, which displaces the articulated object so that the joint in-between the links which must be manipulated can be located in the centre of the robot manipulation space, and the action \texttt{takes\_links\_to\_move}, which grasps the links to be manipulated.
Since links may not be successfully grasped by the robot unless they were in the robot manipulation space, this macro aims at providing a single rule for cases whereby links are not in the right position.
\item 
\texttt{changeAngle\_release} is the composition of \texttt{changeAngle}, which changes the orientation of a link, and \texttt{release\_links}, which releases the currently grasped link.
This macro provides a single rule for cases whereby it is necessary to act on a link and release it afterwards.
\item 
\texttt{grasp\_changeAngle\_release} represents the composition of the following actions: \texttt{take\_links\_to\_move}, \texttt{changeAngle}, and \texttt{release\_links}.
This macro provides a single action for cases when it is necessary to act on a link that was already at the centre of robot manipulation space.
\end{itemize}
 
\begin{figure}[t!]
$$
\begin{array}{ll}
m_{1}\ & \texttt{\{linkToCentral\_take(L1,L2,J1,G1,G2,T)\} :-} \texttt{link(L1)},  \texttt{link(L2)}, \texttt{joint(J1)},\\ 
& \qquad \texttt{gripper(G1)}, \texttt{gripper(G2)}, \texttt{time(T)}, \texttt{free(G1,T)}, \texttt{free(G2,T)},\\
& \qquad \texttt{not in\_centre(J1,T)}, \texttt{not in\_hand(L1,T)}, \texttt{not in\_hand(L2,T)},\\
& \qquad \texttt{L1<>L2}, \texttt{G1<>G2},\texttt{connected(J1,L1)}, \texttt{connected(J1,L2)}\\\\
m_{2}\ & \texttt{\{changeAngle\_release(L1,L2,J1,G1,G2,A1,A2,T)\} :-link(L1),link(L2),}\\
& \qquad \texttt{joint(J1), gripper(G1), gripper(G2), angles(A1), angles(A2),}\\
& \qquad \texttt{in\_centre(J1,T), not free(G1,T), not free(G2,T), in\_hand(L1,T),}\\
& \qquad \texttt{in\_hand(L2,T), connected(J1,L1), connected(J1,L2),}\\
& \qquad \texttt{grasped(G1,L1,T), hasAngle(L1,A2,T), grasped(G2,L2,T),}\\
& \qquad \texttt{L1<>L2, G1<>G2.}\\\\
m_{3}\ & \texttt{\{grasp\_changeAngle\_release(L1,L2,J1,A1,A2,G1,G2,T)\}:- link(L1),}\\
& \qquad  \texttt{link(L2), joint(J1), angles(A1), angles(A2), gripper(G1),}\\
& \qquad \texttt{gripper(G2), time(T), in\_centre(J1,T), free(G1,T), free(G2,T),}\\
& \qquad \texttt{connected(J1,L1), connected(J1,L2), in\_centre(J1,T),}\\
& \qquad  \texttt{hasAngle(L1,A2,T), time(T), L1<>L2, G1<>G2.}
\end{array}
$$
\caption{Macros encoding for the extended scenario.}
\label{fig:extended_m}
\end{figure}

\paragraph{\it Macros Encoding for the Extended Scenario.}
The MAES encoding is an improved version of the SAES encoding, and it is based on principles similar to those employed in the SAS and SAES encodings.
The general structure and the way the MAES encoding deals with robot manipulation actions are the same shown in Figure \ref{fig:standard_f}. 
Figure~\ref{fig:extended_m} partially reports how the three macros introduced above are encoded in ASP, considering only preconditions and choice rules, and not showing the effects.
%
To give an overall idea, \textit{MAES} encoding is composed of these three macros, and their effects plus (modified versions of) the rules from $r_{26}$ to $r_{30}$, which are necessary to ensure that only one action is selected at each time step, the rules from $r_{31}$ to $r_{33}$, and eventually $r_{15}$ that ensure the goal is reached. 
Rules from $r_{26}$ to $r_{33}$ must be modified according to other rules of the encoding, e.g., rules from $r_{26}$ to $r_{29}$ are reduced to just three rules, one for each macro. As already stated, the full MAES encoding is reported and explained in Appendix A.

\section{Validation in the Extended Scenario and Experimental Analysis}
\label{sec:exp}

In this Section we present and discuss the results of the analysis we carried out on the extended scenario. 
We first present an experimental validation of the whole framework in the extended scenario, in a way similar to what we have done in Section~\ref{sec:vali} for the simple scenario, and then we discuss an analysis about the contribution of macros on the \textit{Action Planning} module.


\begin{figure}[t!]
$$
\begin{array}{l}
\texttt{link(1..5).}\quad
\texttt{joint(1..4).}\\
\texttt{\#const granularity = 60.}\quad
\texttt{time(0..timemax).}\\
\texttt{angle(0).}\quad
\texttt{angle(60).}\quad
\texttt{angle(120).}\\
\texttt{angle(180).}\quad
\texttt{angle(240).}\quad
\texttt{angle(300).}\\
\texttt{connected(1,1).}\quad
\texttt{connected(1,2).}\quad
\texttt{connected(2,2).}\quad
\texttt{connected(2,3).}\\
\texttt{connected(3,3).}\quad
\texttt{connected(3,4).}\quad
\texttt{connected(4,5).}\quad
\texttt{connected(4,5).}\\
\texttt{hasAngle(1,0,0).}\quad
\texttt{hasAngle(2,90,0).}\quad
\texttt{hasAngle(3,0,0).}\\
\texttt{hasAngle(4,60,0).}\quad
\texttt{hasAngle(5,120,0).}\\
\texttt{goal(1,0).}\quad
\texttt{goal(2,90).}\quad
\texttt{goal(3,0).}\quad
\texttt{goal(4,300).}\quad
\texttt{goal(5,300).}\\
\end{array}
$$

$$
\begin{array}{ll}
a_{s1}: & \texttt{take\_links\_to\_move(4,3,3,0,1,0)}\\
a_{s2}: & \texttt{changeAngle(4,3,3,0,60,0,1,1)}\\
a_{s3}: & \texttt{changeAngle(4,3,3,300,0,0,1,2)}\\
a_{s4}: & \texttt{release\_links(3,4,3,1,0,3)}\\
a_{s5}: & \texttt{move\_link\_to\_central(4,4,0,4) }\\
a_{s6}: & \texttt{take\_links\_to\_move(5,4,4,1,0,5)}\\
a_{s7}: & \texttt{changeAngle(5,4,4,300,0,1,0,6)}\\
\end{array}
$$

$$
\begin{array}{ll}
a_{m1}: & \texttt{grasp\_changeAngle\_release(4,3,3,0,60,0,1,0)}\\
a_{m2}: & \texttt{grasp\_changeAngle\_release(4,3,3,300,0,0,1,1)}\\
a_{m3}: & \texttt{linkToCentral\_take(5,4,4,4,0,1,2)}\\
a_{m4}: & \texttt{changeAngle\_release(5,4,4,0,1,300,0,3)}\\
\end{array}
$$
\caption{An example of problem instance in the extended scenario: the knowledge base of the problem solved with both SAES and MAES (top); two excerpts of the answer set returned by Clingo, respectively when SAES (middle) or MAES (bottom) are employed.
}
\label{fig:extended_val}
\end{figure}

\subsection{Validation in the Extended Scenario}
\label{sec:valext}

We employed the same experimental setting already described in Section~\ref{sec:vali}. 
Figure \ref{fig:extended_val} shows the knowledge base of one sample problem instance, as well as two excerpts of the answer set produced by Clingo when using the SAES and MAES encoding, respectively. 
From the Figure, it is possible to appreciate the differences in the knowledge base with respect to the one ones in the simple scenario.

We can compare the solutions generated by SAES and MAES: 
in the plan generated by the SAES encoding, 
$a_{s1}$ denotes the fact that the robot must take links $3$ and $4$ with the left and the right grippers, respectively;
in $a_{s2}$, the orientation of link $4$ must change from $60$ to $0$ $deg$;
$a_{s3}$ foresees that the orientation of link $4$ must change from $0$ to $300$ $deg$;
in $a_{s4}$, the two links must be released; 
in $a_{s5}$, joint $4$ (located between links $4$ and $5$) must be placed at the centre of robot manipulation space;
according to $a_{s6}$, the robot must grasp links $4$ and $5$ with the left and the right grippers, respectively;
finally, in $a_{s7}$ the orientation of link $5$ must change from $0$ to $300$ $deg$.
Instead, according to the plan generated using the MAES encoding,
$a_{m1}$ requires the robot to grasp links $3$ and $4$ with the left and the right grippers, respectively, that the orientation of link $4$ be changed from $60$ to $0$ $deg$, and that links are released; 
according to $a_{m2}$, the robot must grasp links $3$ and $4$ with the left and the right grippers, respectively, change the orientation of link $4$ from $0$ to $300$ $deg$, and then release the links;
in $a_{m3}$, the two links must be grasped; 
in $a_{m4}$, the orientation of link $5$ must be changed from $0$ to $300$ $deg$.

While the two solutions share a number of similarities, there are nonetheless few remarkable differences. 
It is possible to notice that with the macros it is not possible to act on a link without releasing it just afterwards: 
this leads to a small unnecessary idle time if the robot has to move a link two times in a row since it has to release it and grasp it again.
We avoid that problem in the SAES encoding since the two actions \texttt{changeAngle} and \texttt{releaseLinks} are not encapsulated in a macro.

\begin{figure}[t!]
\centering
\subfigure{
\includegraphics[width=0.23\columnwidth]{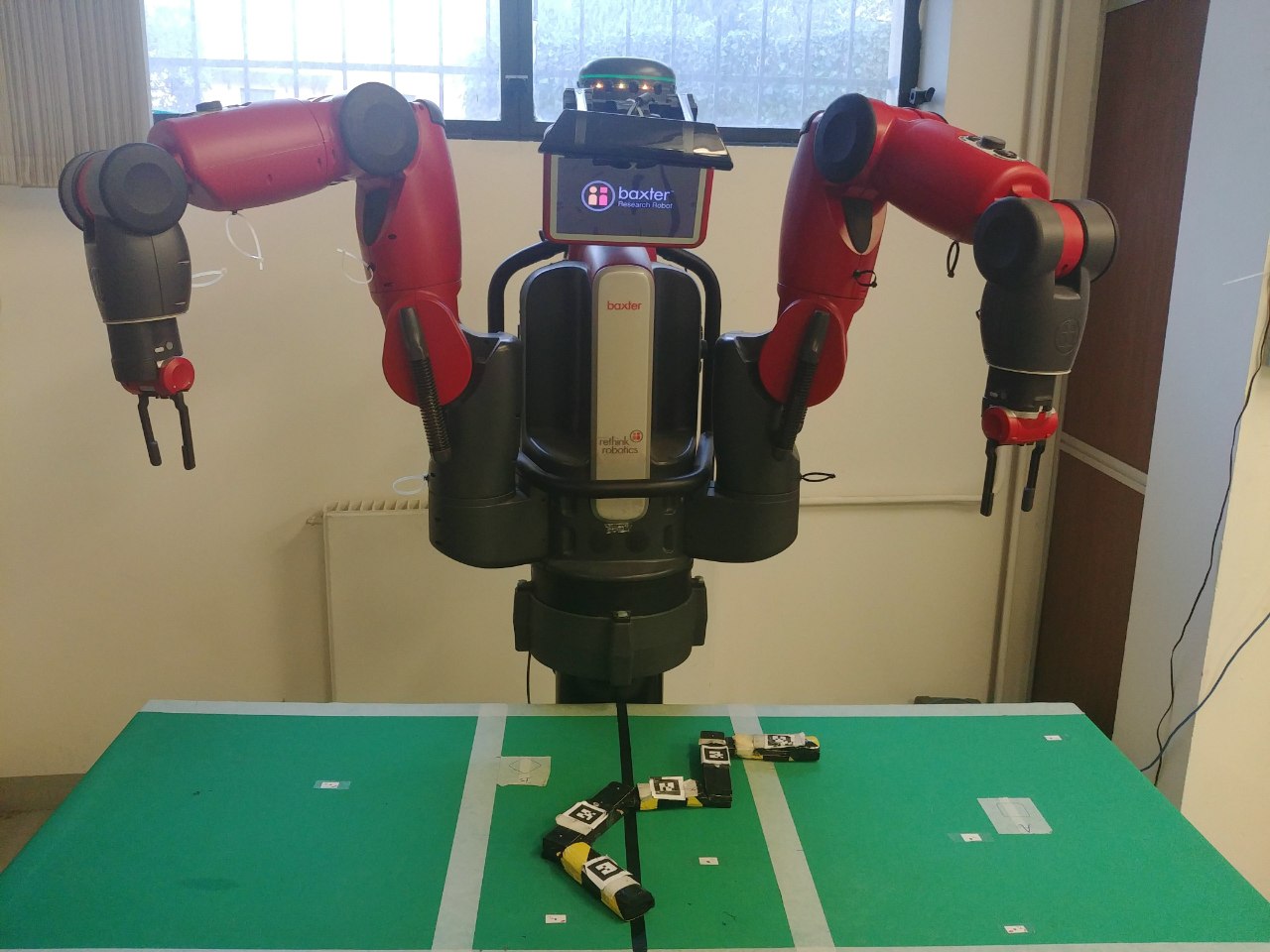}
\label{fig:extended1}}
\subfigure{
\includegraphics[width=0.23\columnwidth]{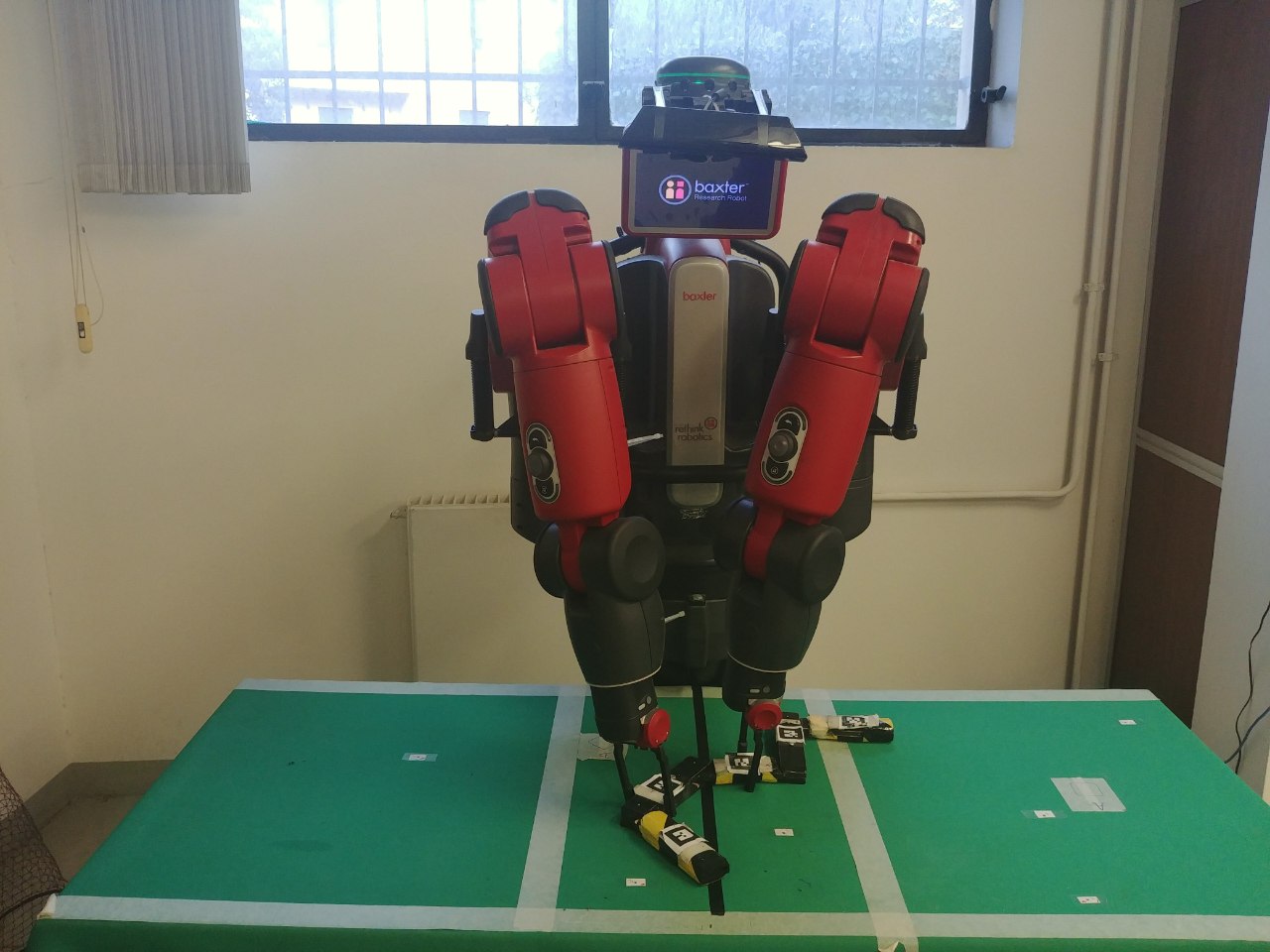}
\label{fig:extended2}}
\subfigure{
\includegraphics[width=0.23\columnwidth]{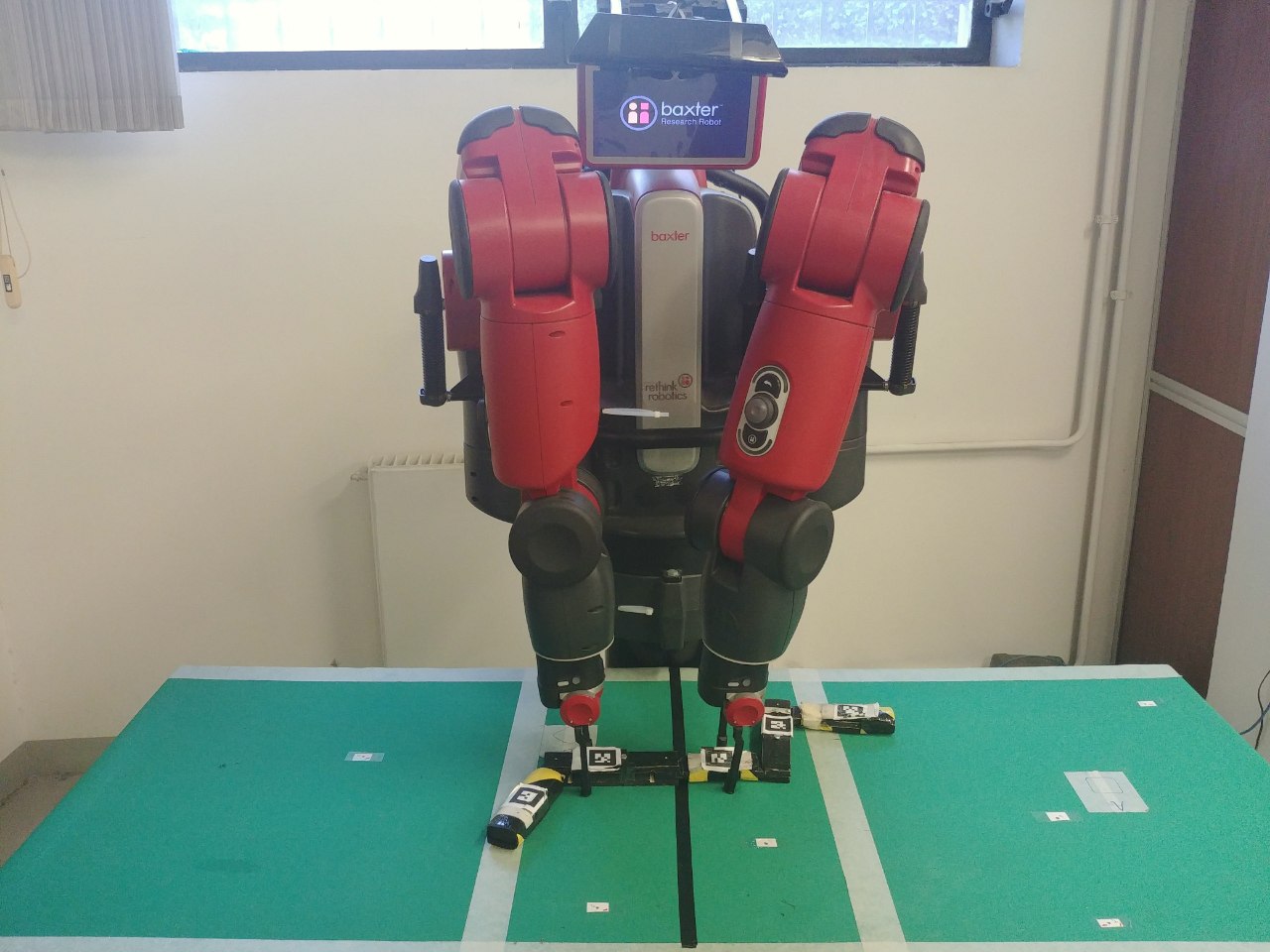}
\label{fig:extended3}}
\subfigure{
\includegraphics[width=0.23\columnwidth]{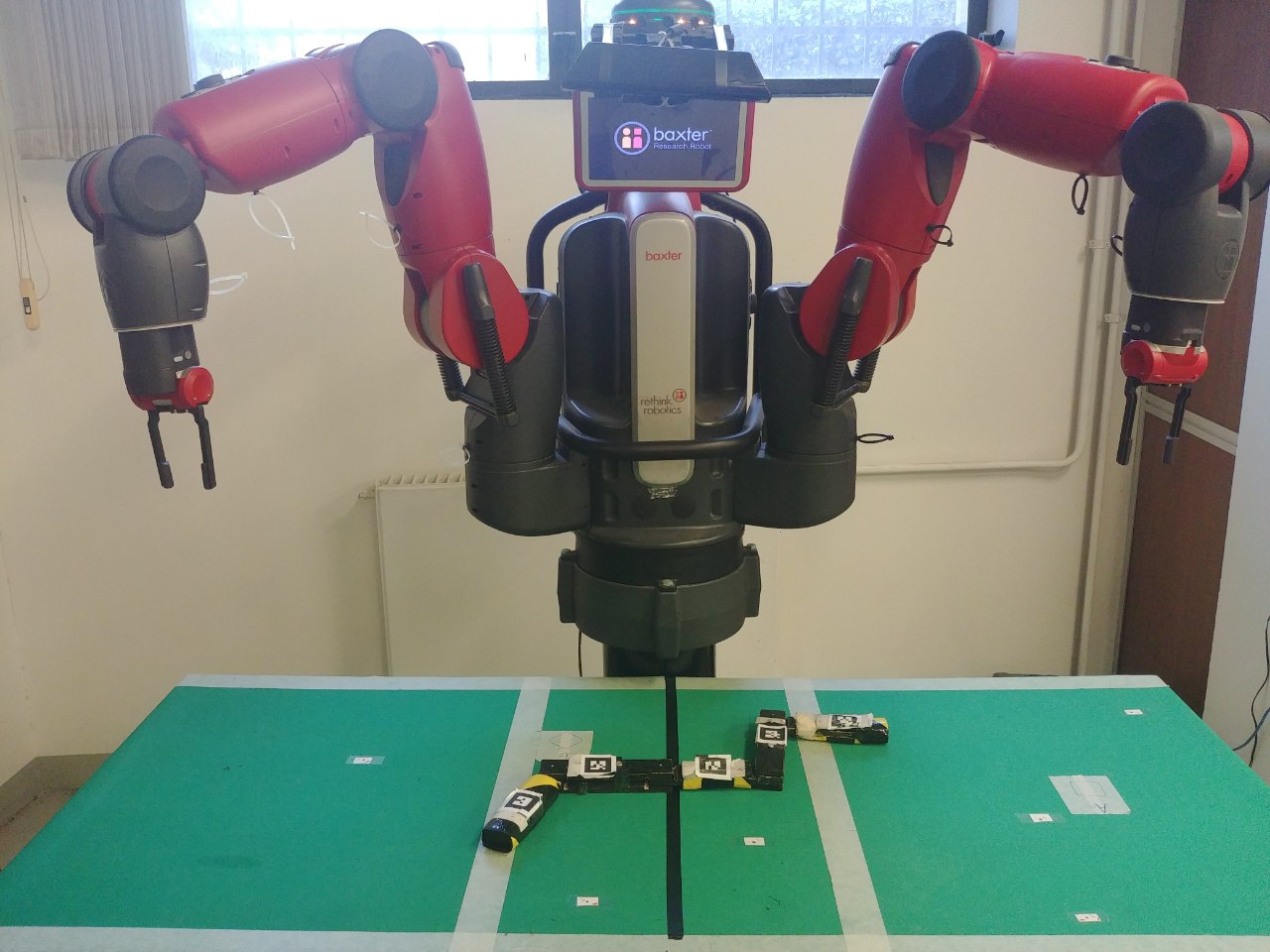}
\label{fig:extended4}}
\subfigure{
\includegraphics[width=0.23\columnwidth]{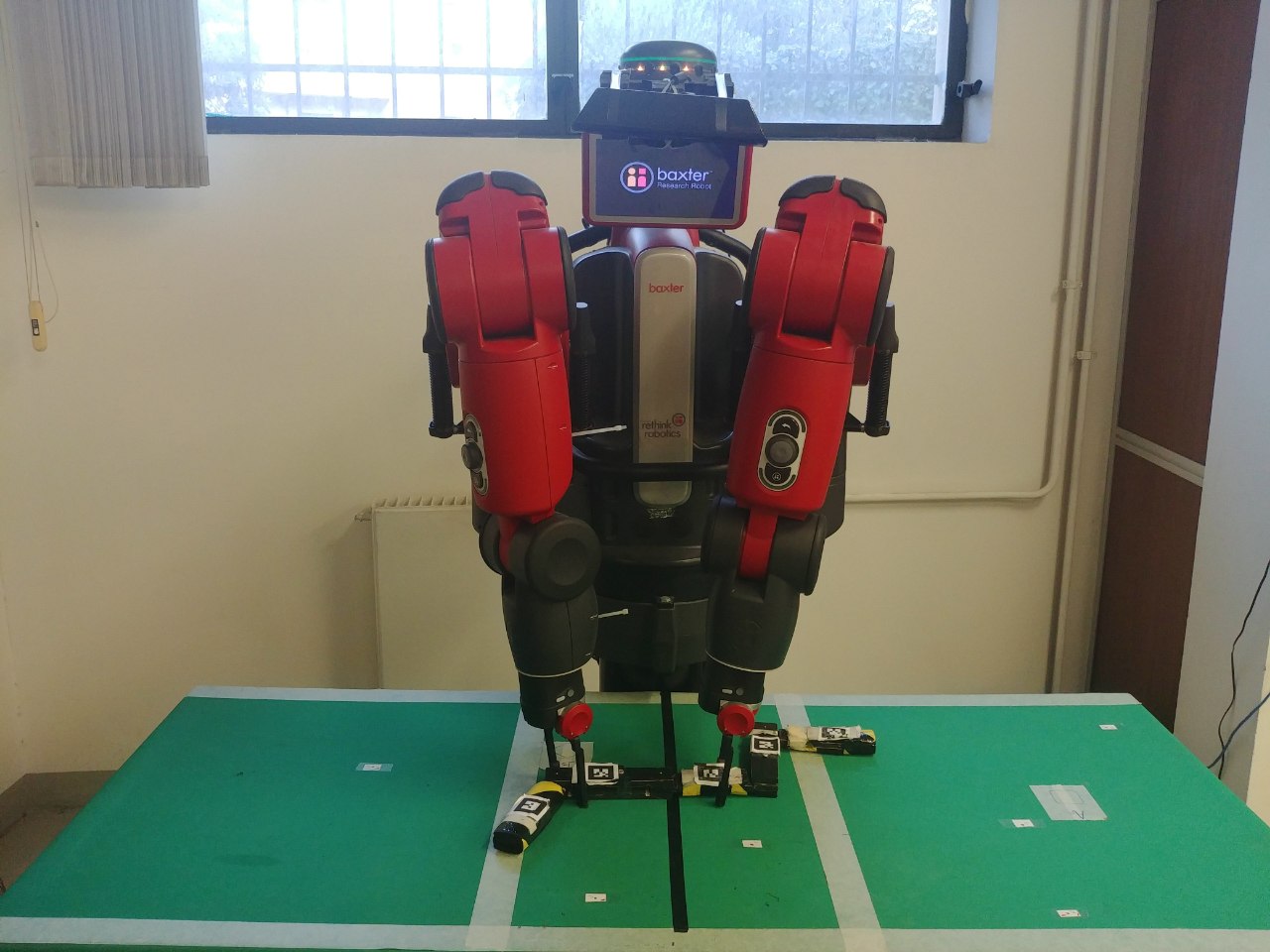}
\label{fig:extended5}}
\subfigure{
\includegraphics[width=0.23\columnwidth]{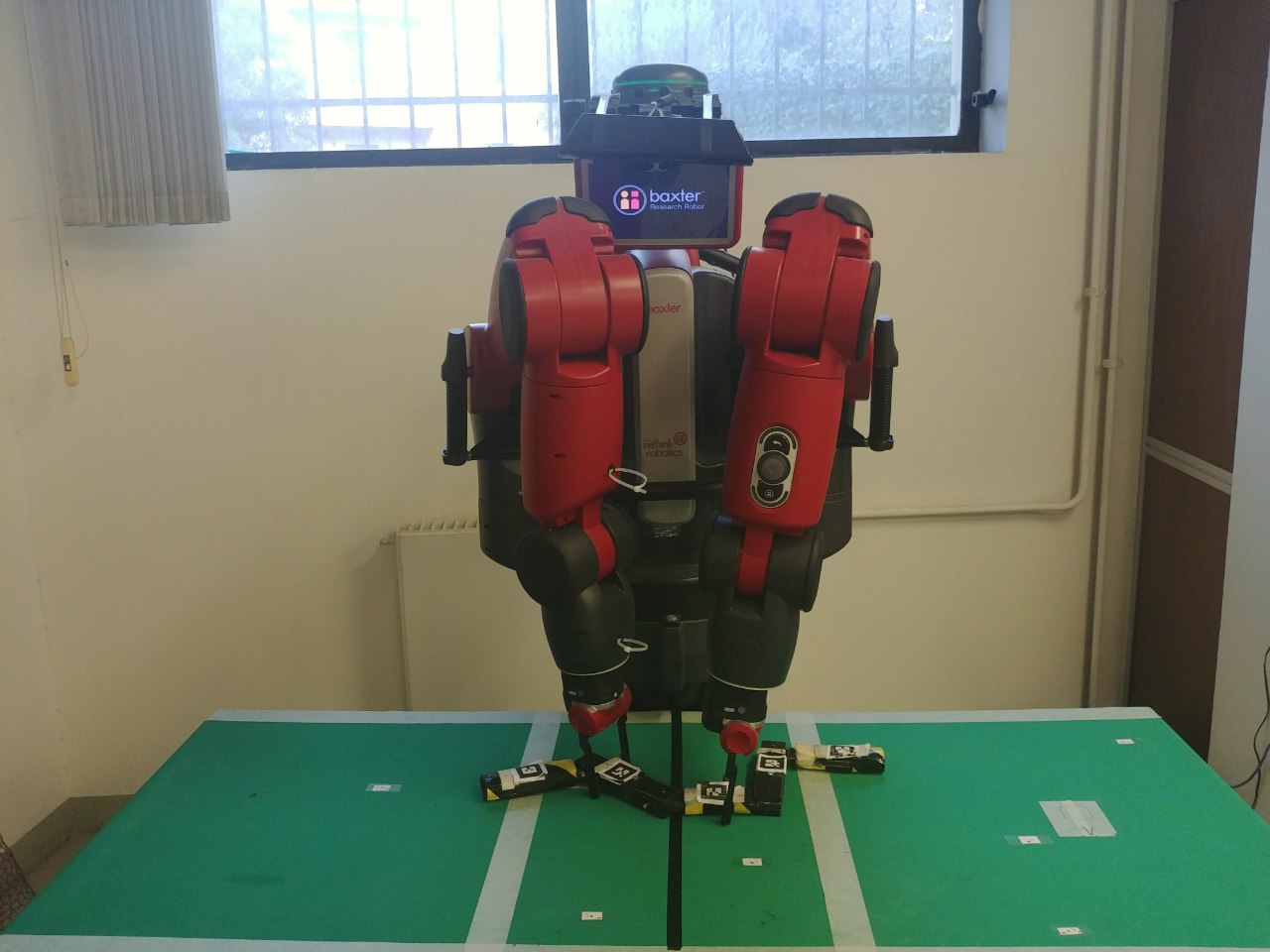}
\label{fig:extended6}}
\subfigure{
\includegraphics[width=0.23\columnwidth]{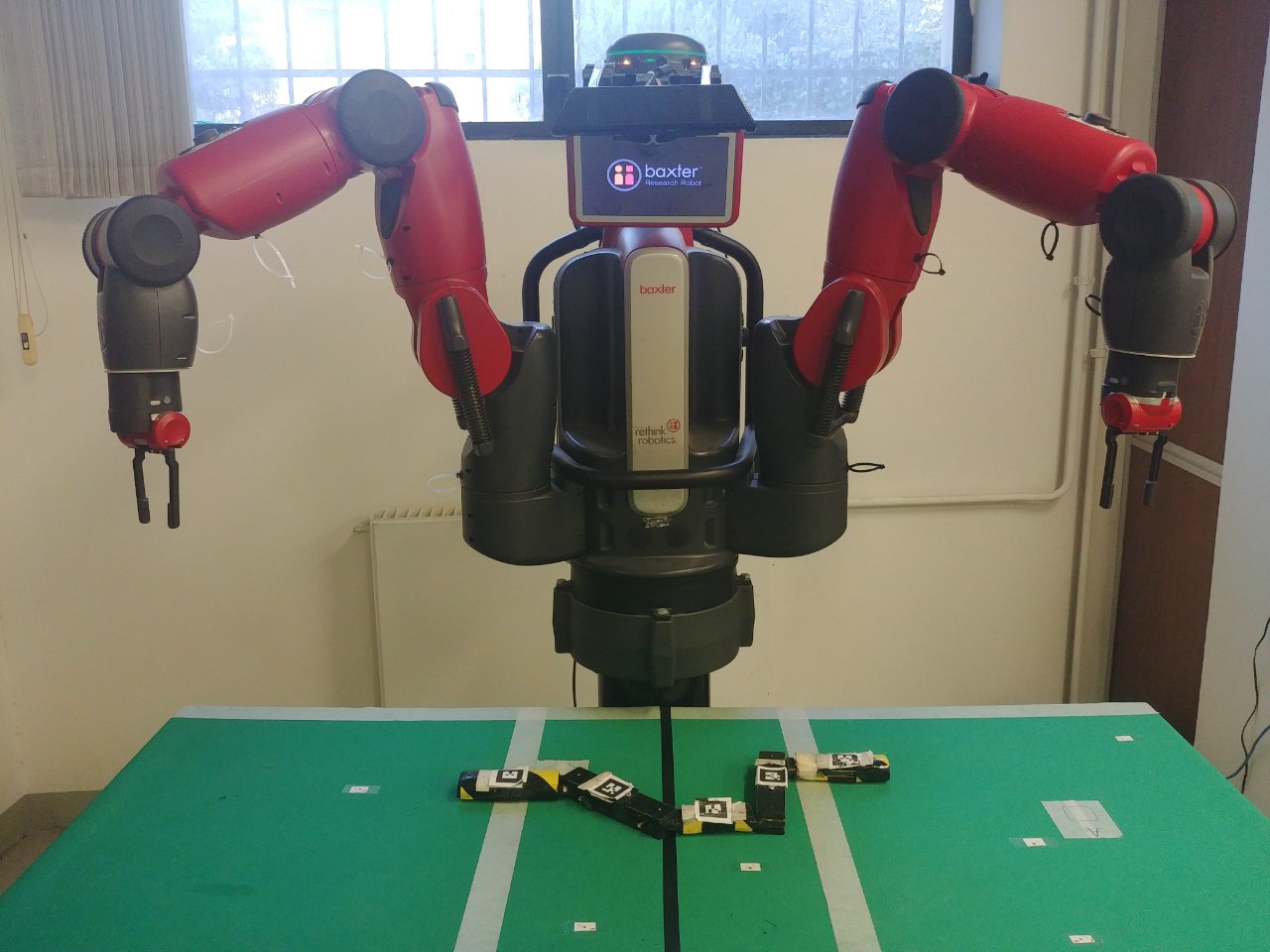}
\label{fig:extended7}}
\subfigure{
\includegraphics[width=0.23\columnwidth]{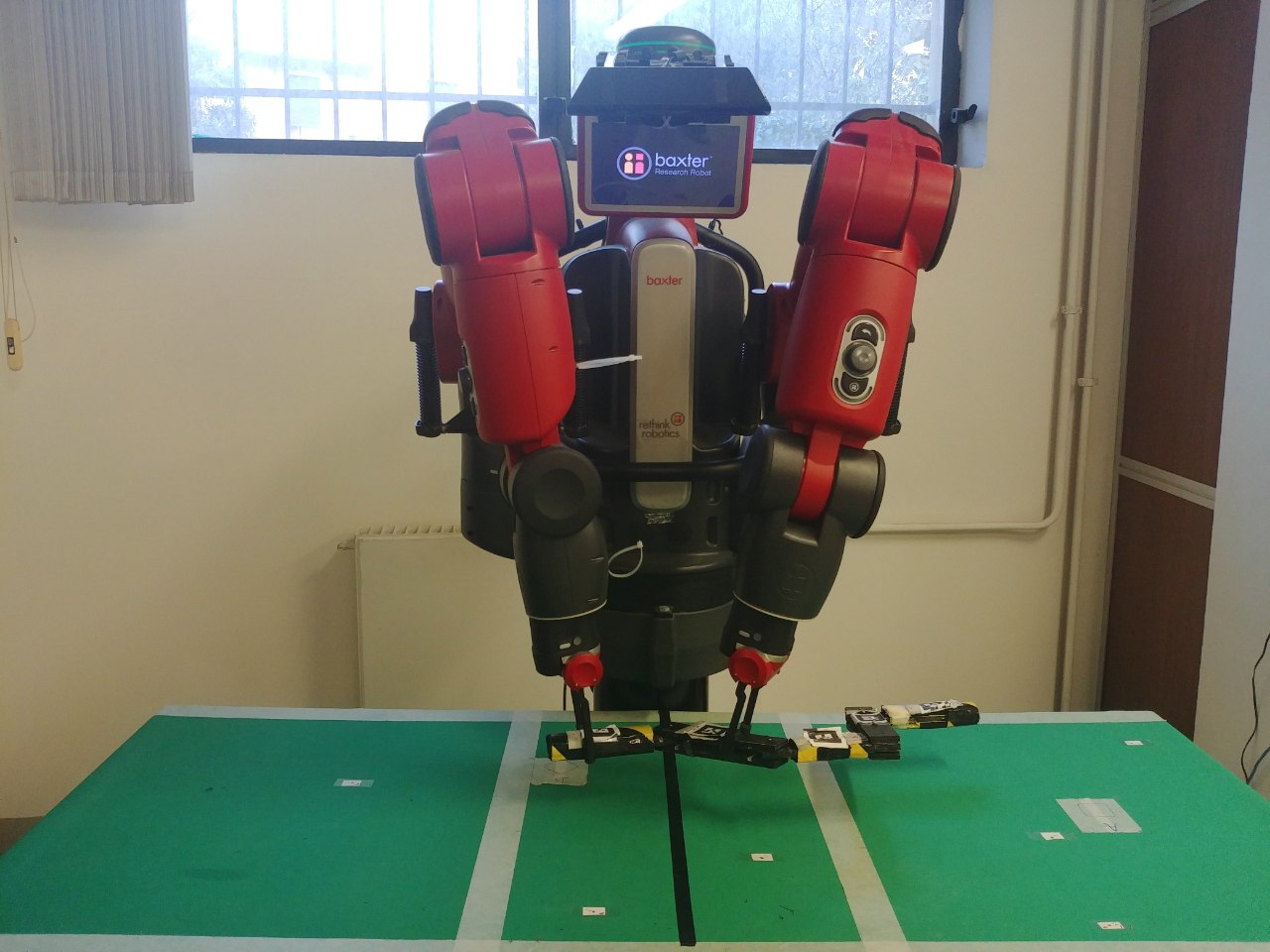}
\label{fig:extended8}}
\subfigure{
\includegraphics[width=0.23\columnwidth]{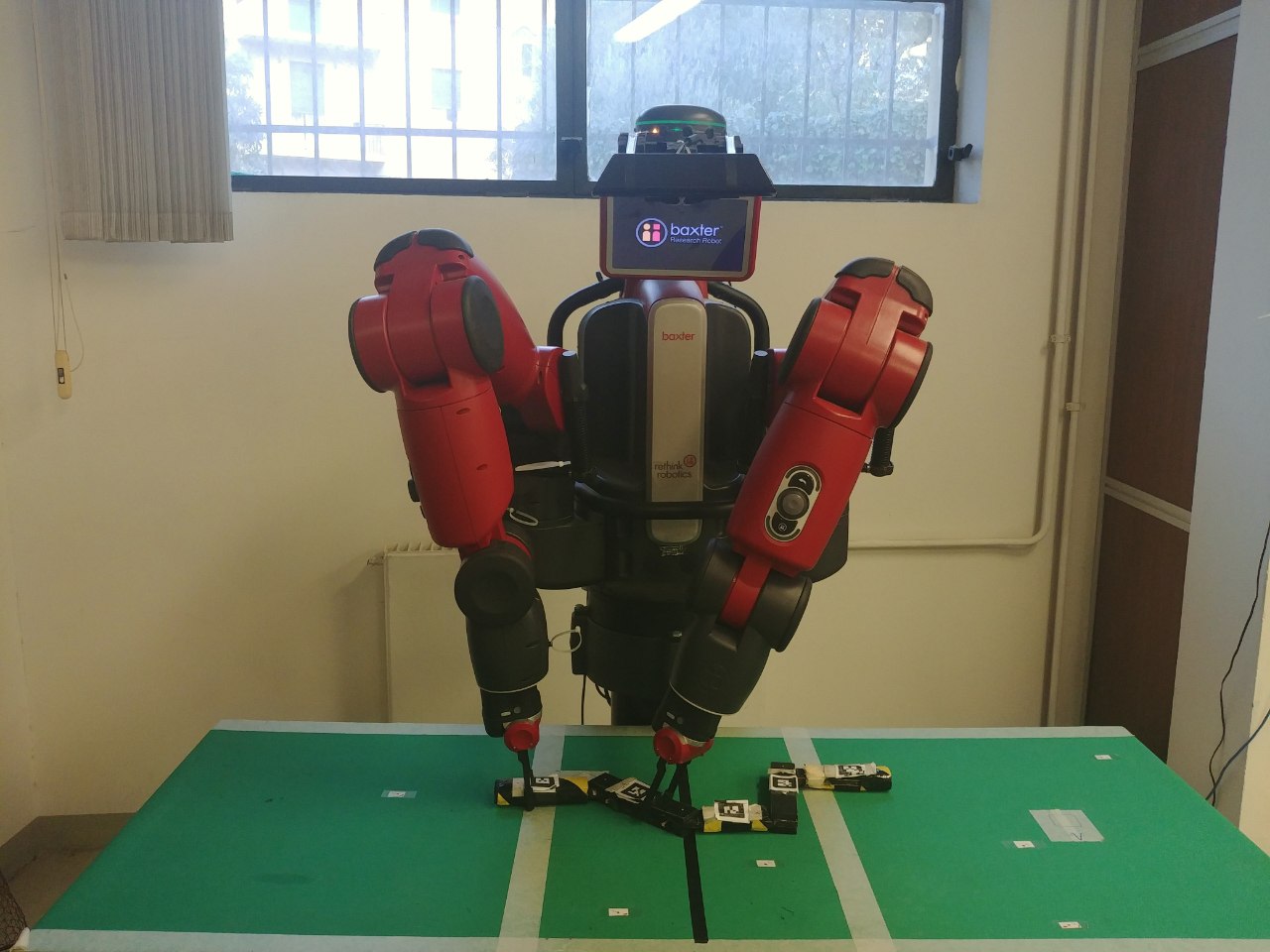}
\label{fig:extended9}}
\subfigure{
\includegraphics[width=0.23\columnwidth]{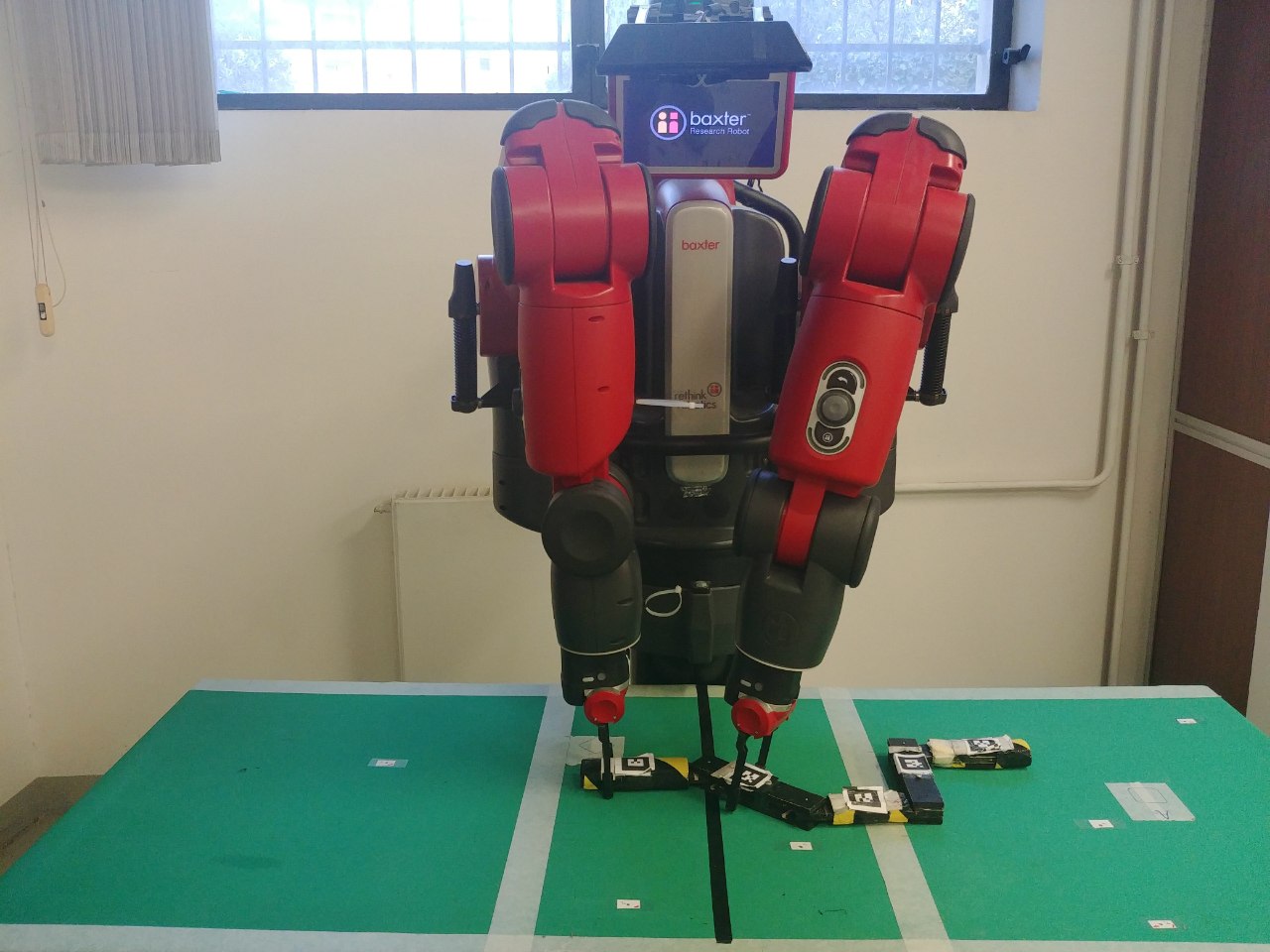}
\label{fig:extended10}}
\subfigure{
\includegraphics[width=0.23\columnwidth]{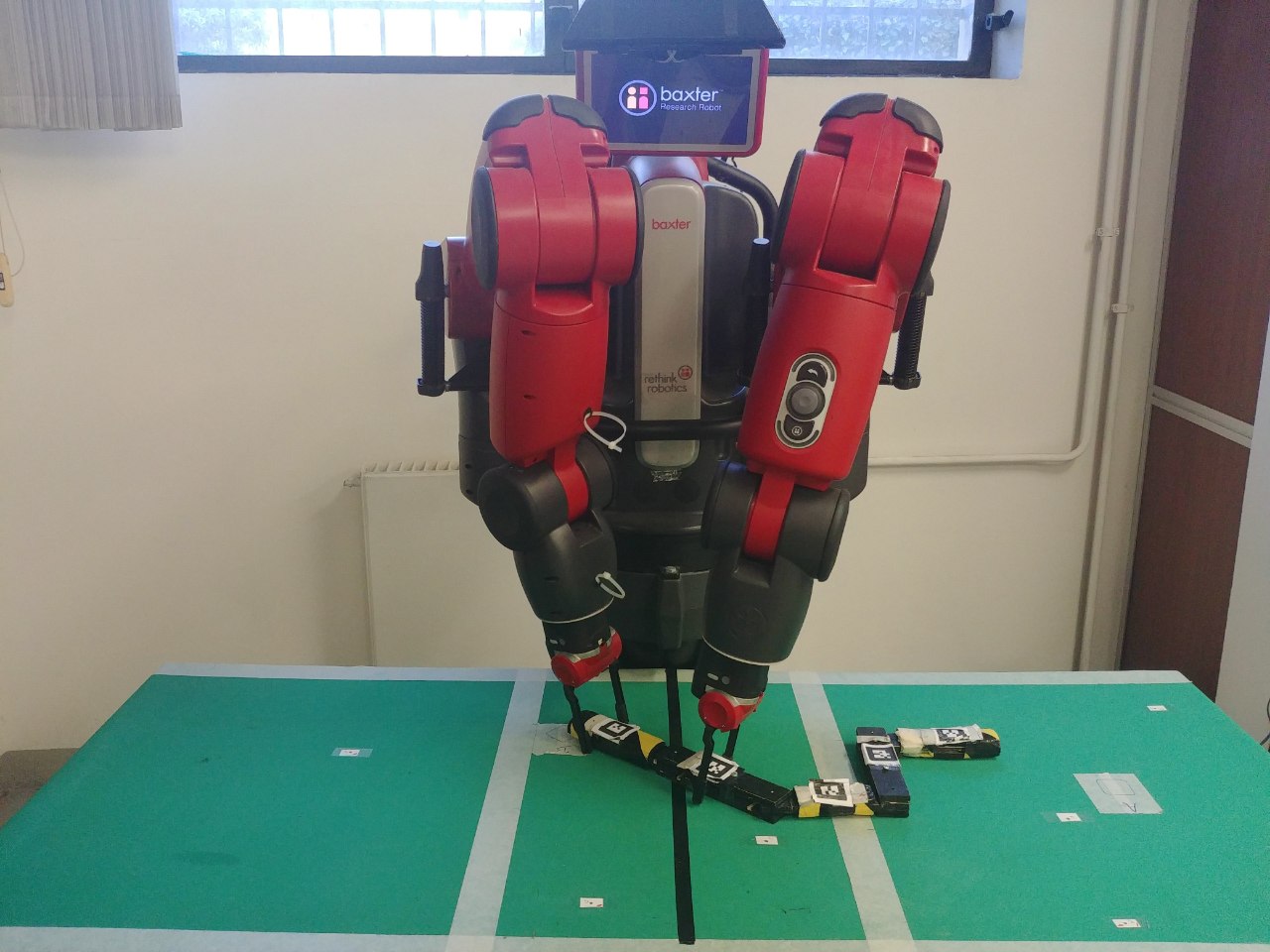}
\label{fig:extended11}}
\subfigure{
\includegraphics[width=0.23\columnwidth]{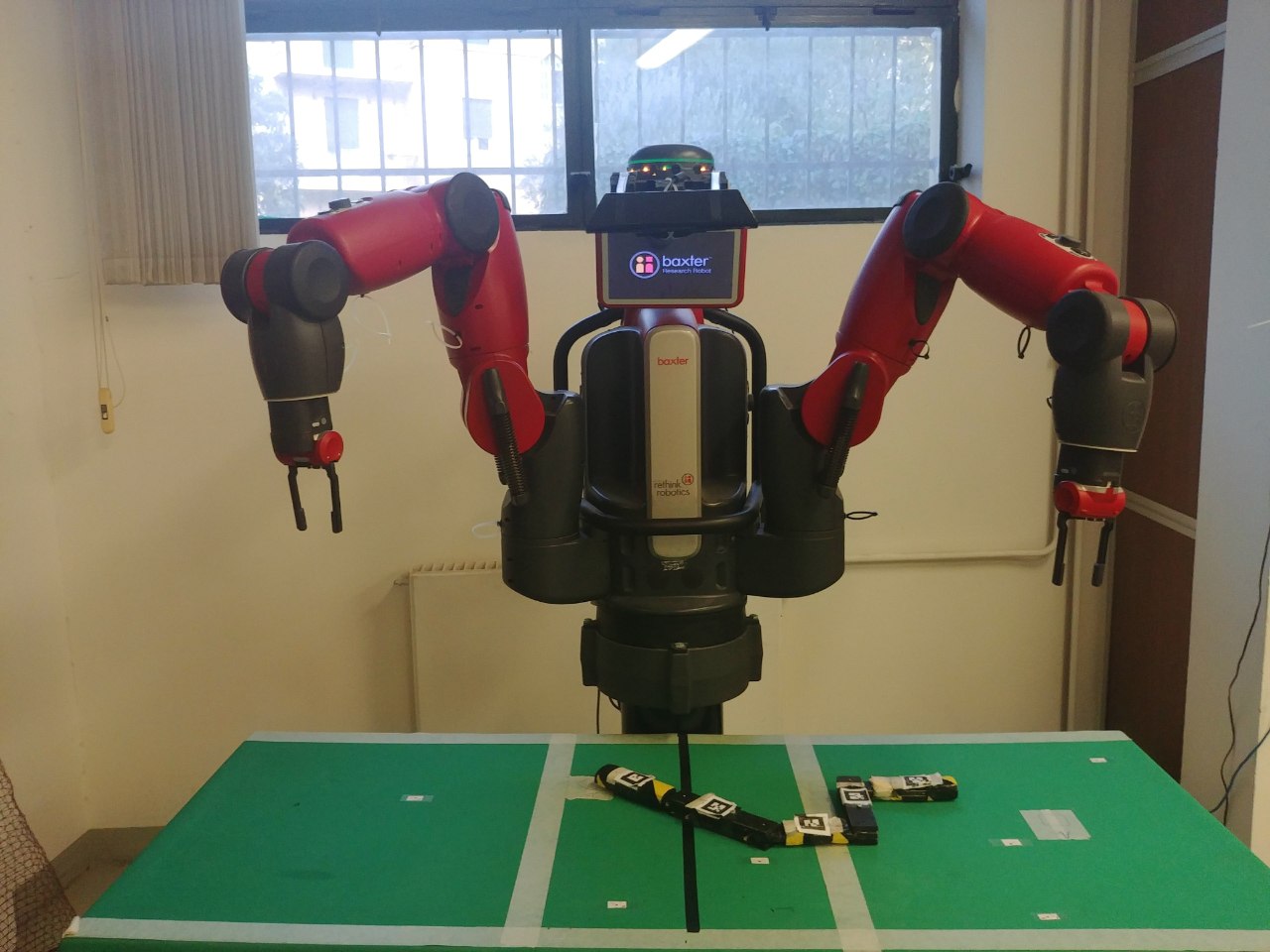}
\label{fig:extended12}}
\caption{Execution of a problem instance in the complex scenario.}
\label{fig:maes_validation}
\end{figure}

Figure \ref{fig:maes_validation} illustrates the execution of the solution generated using the {MAES} encoding (from top-left to bottom-right), on the sample instance at the top of Figure \ref{fig:extended_val}, which produces the (partial) answer set at the bottom of the same Figure. 
Starting from the initial configuration of the articulated object (top-left), the up row represents the three steps associated with macro action $a_{m1}$, i.e., grasping, link angle change, and release, respectively.
Next, the first three Figures in the mid row represent action $a_{m2}$, which is almost equivalent to action $a_{m1}$. 
The last Figure in the mid row and first two Figures of the bottom raw represent three steps of action $a_{m3}$, which brings the fourth joint to the centre of robot manipulation space, whereas the third Figure in the bottom row shows the changing orientation step of  action $a_{m4}$. 
It is noteworthy that even if moving the object towards the centre of the manipulation space strictly requires only one gripper, during action execution we hard-code the use of two grippers to increase precision.  
Finally, the bottom-right Figure shows the final configuration corresponding to the required goal configuration of the $5$-link articulated object.
It is important to notice that, for the sake of brevity, we skipped some of the steps involved in actions $a_{m2}$ and $a_{m3}$. 
It is interesting to notice, however, although the steps shown in Figure \ref{fig:maes_validation} would have been skipped by a {SAES}-generated plan (see Figure \ref{fig:extended_val}), therefore reducing the robot idle time, that the flow of the actions during the execution process of both {MAES} and {SAES} is almost equivalent even if the organisation of the plan is different. However, we will see in the next Subsection the computational advantages that MAES brings.

\subsection{Experimental Analysis on the Action Planning module}
\label{subsec:eaapm}

In order to obtain an assessment of the capabilities of the developed encodings, we selected and applied to the new, extended scenario some of the $280$ instances generated for the simple scenario. 
Eventually, we considered $180$ problem instances with the number of links varying from $4$ to $12$, and with orientation granularity values of $4$, $6$, $8$, or $12$ possible angles, $5$ instances for each pair (number of links, granularity).
For each problem instance, a time limit of $300$ seconds and a memory limit of $16$ GB was applied.
Clingo was used to solve the ASP-encoded instances. 
All the experiments have been conducted on Intel i7-4790 CPU and Linux OS. 

We compared the performance of the considered encodings using \textit{coverage}, i.e., the percentage of solved instances, and the Penalised Average Run-time (PAR10) score.
The latter is a metric usually exploited in machine learning and algorithm configuration techniques.
It trades off coverage and run-time for solved problems: 
if an encoding $e$ allows the solver to solve an instance $\Pi$ in time $t\leq T$ ($T = 300$ $s$ in our case), then $PAR10(e,\Pi) = t$, otherwise $PAR10(e,\Pi) = 10 \times T$ (i.e., $3000s$ in our case).

\begin{table}











\begin{tabular}{c|| c | c c ||c |c c }

\multicolumn{7}{c}{Number of allowed orientations: 4} \tabularnewline
& \multicolumn{3}{c||}{PAR10} & \multicolumn{3}{c}{Coverage} \tabularnewline
\cmidrule{1-7}
& \textbf{SAS} & \textbf{SAES} &\textbf{ MAES} & \textbf{SAS} & \textbf{SAES} & \textbf{MAES} \tabularnewline
\cmidrule{1-7}
\textbf{4} & 0.02 & 2.0 & 0.3 & 100.0 & 100.0 & 100.0 \tabularnewline
\textbf{5} & 0.06 & 622.5 & 6.6 & 100.0 & 80.0 & 100.0 \tabularnewline
\textbf{6} & 615.54 & 1852.7 & 676.3 & 100.0 & 40.0 & 80.0 \tabularnewline
\textbf{7} & 197.31 & 1817.1 & 1505.8 & 80.0 & 40.0 & 40.0 \tabularnewline
\textbf{8} & 18.97 & 2411.9 & 1254.3 & 100.0 & 20.0 & 60.0 \tabularnewline
\textbf{9} & 92.41 & 3000.0 & 3000.0 & 100.0 & -- & -- \tabularnewline
\textbf{10} & 1335.49 & 3000.0 & 3000.0 & 60.0 & -- & -- \tabularnewline
\textbf{11} & 1803.77 & 3000.0 & 3000.0 & 40.0 & -- & -- \tabularnewline
\textbf{12} & 2401.08 & 3000.0 & 3000.0 & 20.0 & -- & -- \tabularnewline
\cmidrule{1-7}
\cmidrule{1-7}
 \multicolumn{7}{c}{Number of allowed orientations: 6} \tabularnewline
\cmidrule{1-7}
& \textbf{SAS} & \textbf{SAES} &\textbf{ MAES} & \textbf{SAS} & \textbf{SAES} & \textbf{MAES} \tabularnewline
\cmidrule{1-7}
\textbf{4} & 0.04 & 35.6 & 16.9 & 100.0 & 100.0 & 100.0 \tabularnewline
\textbf{5} & 2.028 & 1800.1 & 36.3 & 100.0 & 40.0 & 100.0 \tabularnewline
\textbf{6} & 628.962 & 3000.0 & 2400.6 & 80.0 & -- & 20.0 \tabularnewline
\textbf{7} & 637.682 & 3000.0 & 3000.0 & 80.0 & -- & -- \tabularnewline
\textbf{8} & 1816.754 & 3000.0 & 3000.0 & 40.0 & -- & -- \tabularnewline
\textbf{9} & 2718.2 & 3000.0 & 3000.0 & 20.0 & -- & -- \tabularnewline
\cmidrule{1-7}
\cmidrule{1-7}
 \multicolumn{7}{c}{Number of allowed orientations: 8} \tabularnewline
\cmidrule{1-7}
& \textbf{SAS} & \textbf{SAES} &\textbf{ MAES} & \textbf{SAS} & \textbf{SAES} & \textbf{MAES} \tabularnewline
\cmidrule{1-7}
\textbf{4} & 6.762 & 671.7 & 665.6 & 100.0 & 80.0 & 80.0 \tabularnewline
\textbf{5} & 43.816 & 3000.0 & 1860.8 & 100.0 & -- & 40.0 \tabularnewline
\textbf{6} & 1239.81 & 3000.0 & 2454.9 & 60.0 & -- & 20.0 \tabularnewline
\textbf{7} & 1822.898 & 3000.0 & 3000.0 & 40.0 & -- & -- \tabularnewline
\cmidrule{1-7}
\cmidrule{1-7}
 \multicolumn{7}{c}{Number of allowed orientations: 12} \tabularnewline
\cmidrule{1-7}
& \textbf{SAS} & \textbf{SAES} &\textbf{ MAES} & \textbf{SAS} & \textbf{SAES} & \textbf{MAES} \tabularnewline
\cmidrule{1-7}
\textbf{4} & 614.232 & 3000.0 & 101.2 & 80.0 & -- & 80.0 \tabularnewline
\textbf{5} & 1877.986 & 3000.0 & 3000.0 & 40.0 & -- & -- \tabularnewline
\textbf{6} & 1232.116 & 3000.0 & 3000.0 & 40.0 & -- & -- \tabularnewline
\textbf{7} & 2809.50 & 3000.0 & 3000.0 & 20.0 & -- & -- \tabularnewline
\cmidrule{1-7}
\cmidrule{1-7}
\end{tabular}
\caption{
Results, in terms of PAR10 and coverage, achieved by the considered encodings on the tested instances. 
Instances are grouped according to the number of links of the articulated object to be manipulated (rows) and the granularity of the angular values, with $5$ instances for each pair (number of links, granularity). 
Cases not solved by any considered encoding are omitted.}
\label{fig:t_actions}
\end{table}

Table \ref{fig:t_actions} summarises the results achieved by Clingo to solve instances encoded in {SAS}, {SAES}, and {MAES}. 
It is noteworthy to recall that {SAS} is much more simplistic than the other encodings, as it ignores the position of the links to be manipulated, and considers high-level actions which have to be broken down into a large number of low-level elementary actions.
On the contrary, {SAES} and {MAES} provide a more detailed and rich description of the problem, which allows to generating plans easier to be put in place by the robot. 
Therefore, a direct comparison between {SAS} and {SAES/MAES} is not straightforward, but it is nonetheless interesting to analyse also the results obtained by {SAS}.
Unsurprisingly, the {SAS} encoding solves a larger number of instances, and is generally the fastest among the considered encodings. 
Due to the above-mentioned issues, the excellent planning performance of {SAS} comes at the cost of a potentially large number of failures in execution, since the encoding does not take into consideration the effective boundaries of the robot workspace.
During the execution of the computed plans the robot may have to perform actions on joints which are at the limit of the robot manipulation space, therefore causing the robot arms to singularity positions and to stop plan execution.
It is trivial to notice that the more links the articulated object has, the easier it is to experience such failures.

The comparison of the performance achieved by Clingo when using {SAES} and {MAES} can shed some light on the usefulness of macros. 
It can be noticed that macros allow Clingo to solve a larger number of instances, and that macros are generally helpful in improving the run-time. 
This is true regardless of the granularity considered for allowed link orientations, and of the number of considered links. 
As an example, with a granularity value of $12$ (i.e., joint's angles can be modified by $30$ $deg$ per movement), the use of macros allow Clingo to solve $80\%$ of instances of size $4$; no instances can be solved using {SAES}, instead.




In summary, it is evident from Table~\ref{fig:t_actions} that macros improve considerably planning performance as well as run-time success rate. 
The introduction of macros leads to better run-time performance, but at the cost of reintroducing idle robot time in certain specific cases.
This is aligned with the results achieved in automated planning when exploiting macros: run-time and coverage performance of planners tend to improve, but the quality of generated plans may be negatively affected due to repetitions and sub-optimalities \cite{ChrpaV19}. 

\section{Related Work}
\label{sec:rela}


In Sections \ref{sec:action} and \ref{sub_asp_modules} we have shown two of our encodings of the Action Planning module, for the simple and extended scenario, respectively, with one particular manipulation mode and search strategy. In Appendix A we provide the encoding exploiting macro actions, with the same manipulation mode and search strategy.
However, as we already stated, we have designed a series of encodings, 
including different search strategies.
In particular, after imposing a reasonable \texttt{timemax}, we have devised further encodings implementing 
$(i)$ a first strategy based on the algorithm \textit{optsat}~\cite{GiunchigliaM06,RosaGM08,DBLP:journals/constraints/RosaGM10}, whereby the heuristic of the solver is modified so as to prefer plans with increasing length, and
$(ii)$ a second strategy employing a choice rule to select the timestep, of course possibly losing optimality (see also \cite{DBLP:conf/lpnmr/DimopoulosGL0S17}).

In \cite{DBLP:conf/aiia/CapitanelliMMV17,DBLP:journals/ras/CapitanelliMMV18} a similar framework based on automated reasoning methodologies was presented. 
The framework employs a Description Logic (DL) based knowledge representation framework and the Planning Domain Definition Language (PDDL), as well as automated planning engines for the \textit{Action Planning} module, whereas we use an ASP-based uniform language and approach for the whole framework, leaving to the Motion Planner module handling the kinematics problems.
Moreover, differently from most of our approaches, models and solvers employed in \cite{DBLP:conf/aiia/CapitanelliMMV17,DBLP:journals/ras/CapitanelliMMV18} are not guaranteed to generate the shortest-length plans, which is indeed important, given that in this context action execution is quite time consuming.

The ASP-based architecture described in this paper can be integrated with ROSoClingo~\cite{DBLP:conf/lpnmr/AndresRSS15}, which is a framework combining the ASP solver Clingo (version~4) with the ROS middleware. 
In particular, it provides a high-level ASP-based interface to control the behaviour of a robot and to process the results of actions execution.
In our framework, the interaction with ROS is handled by a software adapter developed ad-hoc. 
Furthermore, our architecture can be integrated with telingo~\cite{DBLP:conf/lpnmr/CabalarKMS19}, an extension of the ASP system Clingo with temporal operators over finite linear time, which automatically uses the multi-shot interface of Clingo, and therefore it might be useful to further simplify our architecture and improve performance. 

ASP has been employed in different domains, including Robotics~\cite{DBLP:conf/lpnmr/GebserJKOSSS13,DBLP:journals/tplp/ErdemPSSU13,DBLP:conf/lpnmr/AndresRSS15,DBLP:conf/icra/ErdemPS15,DBLP:conf/aips/SchapersNLGS18}, encompassing various application domain. 
However, in the current literature the goal is not the validation and exploitation of the techniques on a real robot, as it is in our case. 
For a recent overview, the interested reader is referred to~\cite{DBLP:journals/ki/ErdemP18}.

Focusing on planning encodings, recently the Plasp system~\cite{DBLP:conf/lpnmr/DimopoulosGL0S17} has been further extended with both SAT-inspired and genuine encodings. 
Some of them have helped reduce the (still existing) gap with automated planning techniques. 
Our aim in the design of the encoding for our scenario is to obtain a working solution for the problem at hand, rather than achieving the best performing solution.
Nonetheless, results discussed in~\cite{DBLP:conf/lpnmr/DimopoulosGL0S17} could be employed to further speed-up our \textit{Action Planning} module.

An important line of research in action planning focuses on increasing the efficiency of the planning process by reformulating the domain knowledge, with the aim of obtaining models which are more amenable for automated reasoners. 
Significant work has been done in the area of reformulation for improving the performance of domain-independent planners. 
\textit{Macro-operators} \cite{mo,ma,my_ker,McCPort} are one of the best known types of reformulation in classical planning.
They encapsulate in a single planning operator a sequence of \textit{original} operators. From a technical perspective, an instance of a macro is applicable in a state if and only if a corresponding sequence of operator instances is applicable in that state and the result of the application of the macro instance is the same as the result of applying the corresponding sequence of operator instances. 
The notion of macros can also be exploited by specifically enhanced planning reasoners. This is the case for MacroFF SOL-EP version~\cite{macroff}, which is able to exploit offline extracted and ranked macros, and Marvin~\cite{ma2}, which generates macros online
by combining sequences of actions previously used for escaping plateaus.
Such systems can efficiently deal with drawbacks of specific planning engines, in this case the FF planner~\cite{ff}.
However, their adaptability for different planning engines might be low.
In this work, we aimed at exploiting macros in ASP following the more traditional solver-independent approach, i.e., by modifying the encoding, replacing simple actions with macros. 

\section{Conclusions}
\label{sec:conc}

In this paper we presented an ASP-based framework for the automated manipulation of articulated objects in a 2D workspace by a dual-arm robot manipulator. 
We demonstrated the validity and usefulness of the proposed approach by running real-world experiments with a real robot in two scenarios.
In the second, extended scenario, we applied to ASP the concept of macros in order to deal with plan generation and execution more efficiently.
Our analysis shows the effectiveness of the proposed ASP-based approach, using Clingo as a solver, and the usefulness of employing macros.

We see several directions for future work. 
First, we are interested in validating the framework on different dual-arm robots, possibly manipulating different articulated objects. 
Given the nature of the approach, we expect it to generalise with a reasonably limited effort. 
We also plan to integrate telingo with our framework.
Then, we plan to develop a mixed encoding composed by simple and macro actions. 
Furthermore, our instances could be an interesting benchmark domain for ASP competitions, e.g.,~\cite{GebserMR17a,GebserMR20}. 
Finally, we would like to evaluate our encodings also with other solvers, e.g., \cite{AlvianoADLMR19} and, given that the encodings contain a significant number of arithmetic operations, to explore the possibility of employing CASP encodings and solvers, e.g., Clingcon~\cite{DBLP:journals/tplp/BanbaraKOS17}, EZCSP~\cite{DBLP:journals/tplp/BalducciniL17} and EZSMT~\cite{DBLP:conf/kr/ShenL18}, to deal with numerical constraints.

All the material presented in this paper, including encodings, instances, results and validations, can be found at \url{https://tinyurl.com/ycbp798j}.

\bibliography{bibtex}
\bibliographystyle{acmtrans}
\newpage
\section*{Appendix A}
\label{ape:a}

\begin{figure}[th!]
$$
\footnotesize
\begin{array}{ll}
m_{1}\ & \texttt{\{linkToCentral\_take(L1,L2,J1,G1,G2,T)\} :-} \texttt{link(L1)}, \texttt{link(L2)},\texttt{joint(J1)},\\ 
& \qquad \texttt{gripper(G1)},\texttt{gripper(G2)},\texttt{time(T)},\texttt{free(G1,T)},\texttt{free(G2,T)},\\
& \qquad \texttt{not in\_centre(J1,T)},\texttt{not in\_hand(L1,T)},\texttt{not in\_hand(L2,T)},\\
& \qquad \texttt{L1<>L2},\texttt{G1<>G2},\texttt{connected(J1,L1)},\texttt{connected(J1,L2)}\\
r_{m1}\ & \texttt{in\_centre(J1,T+1):-linkToCentral\_Grasp(\_,\_,\_,J1,\_,T),T<timemax+1.}\\
r_{m2}\ & \texttt{in\_hand(L,T+1):-linkToCentral\_Grasp(L,\_,\_,\_,\_,T),T<timemax+1.}\\
r_{m3}\ & \texttt{in\_hand(L,T+1):-linkToCentral\_Grasp(\_,L,\_,\_,\_,T),T<timemax+1.}\\
r_{m4}\ & \texttt{grasped(G,L,T+1):-linkToCentral\_Grasp(L,\_,\_,G,\_,T),T<timemax+1.}\\
r_{m5}\ & \texttt{grasped(G,L,T+1):-linkToCentral\_Grasp(\_,L,\_,\_,G,T),T<timemax+1.}\\\\

m_{2}\ & \texttt{\{changeAngle\_release(L1,L2,J1,G1,G2,A1,A2,T)1\}:-link(L1),link(L2),}\\
& \qquad \texttt{joint(J1),gripper(G1),gripper(G2),angles(A1),angles(A2),}\\ 
& \qquad \texttt{in\_centre(J1,T),not free(G1,T),not free(G2,T),L1<>L2,G1<>G2,}\\
& \qquad \texttt{connected(J1,L1),connected(J1,L2),grasped(G1,L1,T),}\\
& \qquad \texttt{in\_hand(L1,T),in\_hand(L2,T),hasAngle(L1,A2,T),}\\
& \qquad \texttt{grasped(G2,L2,T).}\\
r_{m6}\ & \texttt{ok(L1,L2,A,Ai,T):-changeAngle\_release(L1,L2,\_,\_,\_,A,Ai,T),}\\
& \qquad \texttt{F1=(A+granularity)$\setminus$360,F2=(Ai$\setminus$360),F1 = F2,A < Ai.}\\
r_{m7}\ & \texttt{ok(L1,L2,A,Ai,T):-changeAngle\_release(L1,L2,\_,\_,\_,A,Ai,T),}\\
& \qquad \texttt{F1=(Ai+granularity)$\setminus$360,F2=(A$\setminus$360),F1 = F2,A > Ai.}\\
r_{m8}\ & \texttt{ok(L1,L2,A,0,T) :-changeAngle\_release(L1,L2,\_,\_,\_,A,Ai,T),}\\
& \qquad \texttt{A = 360-granularity.}\\
r_{m9}\ & \texttt{ok(L1,L2,0,A,T) :-changeAngle\_release(L1,L2,\_,\_,\_,A,Ai,T),}\\
& \qquad \texttt{A = 360-granularity.}\\
r_{m10}\ & \texttt{:- changeAngle\_release(L1,L2,\_,\_,\_,A,Ai,T),not ok(L1,L2,A,Ai,T).}\\
r_{m11}\ & \texttt{affected(L,An,Ac,T):-changeAngle\_release(L1,L2,\_,\_,\_,A,Ap,T),}\\
& \qquad \texttt{angles(An),hasAngle(L,Ac,T),An = |(Ac + (A-Ap)) + 360|$\setminus$360,}\\
& \qquad \texttt{L>L1,L1>L2.}\\
r_{m12}\ & \texttt{affected(L,An,Ac,T):-changeAngle\_release(L1,L2,\_,\_,\_,A,Ap,T),}\\
& \qquad \texttt{angles(An),hasAngle(L,Ac,T),An = |(Ac + (A-Ap)) + 360|$\setminus$360,}\\
& \qquad \texttt{L<L1,L1<L2.}\\
r_{m13}\ & \texttt{hasAngle(L1,A1,T+1):-changeAngle\_release(L1,L2,\_,\_,\_,A1,\_,T),}\\
& \qquad  \texttt{ T<timemax+1.}\\
r_{m14}\ & \texttt{free(G,T+1):-changeAngle\_release(\_,\_,\_,G,\_,\_,\_,T),T<timemax+1.}\\
r_{m15}\ & \texttt{free(G,T+1):-changeAngle\_release(\_,\_,\_,\_,G,\_,\_,T),T<timemax+1.}\\\\
\end{array}
$$
\caption{Macro Action Extended Scenario (MAES) encoding (Part 1).}
\label{fig:extended_ta}
\end{figure}

\begin{figure}[t!]
\footnotesize
$$
\begin{array}{ll}
m_{3}\ & \texttt{\{grasp\_changeAngle\_release(L1,L2,J1,A1,A2,G1,G2,T)\}:- link(L1),}\\
& \qquad  \texttt{link(L2),joint(J1),angles(A1),angles(A2),gripper(G1),}\\
& \qquad \texttt{gripper(G2),time(T),in\_centre(J1,T),free(G1,T),free(G2,T),}\\
& \qquad \texttt{connected(J1,L1),connected(J1,L2),in\_centre(J1,T),}\\
& \qquad  \texttt{hasAngle(L1,A2,T),time(T),L1<>L2,G1<>G2.}\\
r_{m16}\ & \texttt{ok(L1,L2,A,Ai,T):-grasp\_changeAngle\_release(L1,L2,\_,A,Ai,\_,\_,T),}\\
& \qquad  \texttt{F1=(A+granularity)$\setminus$360,F2=(Ai$\setminus$360),F1 = F2,A < Ai.}\\
r_{m17}\ & \texttt{ok(L1,L2,A,Ai,T):-grasp\_changeAngle\_release(L1,L2,\_,A,Ai,\_,\_,T),}\\
& \qquad  \texttt{F1=(Ai+granularity)$\setminus$360,F2=(A$\setminus$360),F1 = F2,A > Ai.}\\
r_{m18}\ & \texttt{ok(L1,L2,A,0,T) :-grasp\_changeAngle\_release(L1,L2,\_,A,Ai,\_,\_,T),}\\
& \qquad  \texttt{A = 360-granularity.}\\
r_{m19}\ & \texttt{ok(L1,L2,0,A,T) :-grasp\_changeAngle\_release(L1,L2,\_,A,Ai,\_,\_,T),}\\
& \qquad  \texttt{A = 360-granularity.}\\
r_{m20}\ & \texttt{:- grasp\_changeAngle\_release(L1,L2,\_,A,Ai,\_,\_,T),}\\
& \qquad  \texttt{not ok(L1,L2,A,Ai,T).}\\
r_{m21}\ & \texttt{affected(L,An,Ac,T):-grasp\_changeAngle\_release(L1,L2,\_,A,Ap,\_,\_,T),}\\
& \qquad  \texttt{angles(An),hasAngle(L,Ac,T),An = |(Ac + (A-Ap)) + 360|$\setminus$360,}\\
& \qquad  \texttt{L>L1,L1>L2.}\\
r_{m22}\ & \texttt{affected(L,An,Ac,T):-grasp\_changeAngle\_release(L1,L2,\_,A,Ap,\_,\_,T),}\\ 
& \qquad  \texttt{angles(An),hasAngle(L,Ac,T),An = |(Ac + (A-Ap)) + 360|$\setminus$360,}\\
& \qquad  \texttt{L<L1,L1<L2.}\\
r_{m23}\ & \texttt{hasAngle(L1,A1,T+1):-grasp\_changeAngle\_release(L1,L2,\_,A1,\_,\_,\_,T),}\\
& \qquad  \texttt{T<timemax+1.}\\
r_{m24}\ & \texttt{free(G,T+1):-grasp\_changeAngle\_release(\_,\_,\_,\_,\_,G,\_,T),T<timemax+1.}\\
r_{m25}\ & \texttt{free(G,T+1):-grasp\_changeAngle\_release(\_,\_,\_,\_,\_,\_,G,T),T<timemax+1.}\\\\
r_{m26}\ & \texttt{action(T, linkToCentral\_Grasp(L1,L2,J1,G1,G2,T)) :- }\\
& \qquad  \texttt{linkToCentral\_Grasp(L1,L2,J1,G1,G2,T).}\\
r_{m27}\ & \texttt{action(T, changeAngle\_release(L1,L2,J1,G1,G2,A1,A2,T)) :- }\\
& \qquad  \texttt{changeAngle\_release(L1,L2,J1,G1,G2,A1,A2,T).}\\
r_{m28}\ & \texttt{action(T, grasp\_changeAngle\_release(L1,L2,J1,A1,A2,G1,G2,T))}\\
& \qquad  \texttt{:- grasp\_changeAngle\_release(L1,L2,J1,A1,A2,G1,G2,T).}\\
r_{m29}\ & \texttt{:- time(T), \#count\{Z : action(T,Z)\} != 1.}\\
r_{m30}\ & \texttt{free(G,T+1) :- free(G,T), not linkToCentral\_Grasp(\_,\_,\_,G,\_,T),}\\
& \qquad  \texttt{ not linkToCentral\_Grasp(\_,\_,\_,\_,G,T), T < timemax+1.}\\
r_{m31}\ & \texttt{in\_hand(L,T+1) :- in\_hand(L,T),}\\
& \qquad  \texttt{not changeAngle\_release(L,\_,\_,\_,\_,\_,\_,T),}\\
& \qquad  \texttt{not changeAngle\_release(\_,L,\_,\_,\_,\_,\_,T),}\\
& \qquad  \texttt{not grasp\_changeAngle\_release(L,\_,\_,\_,\_,\_,\_,T),}\\
& \qquad  \texttt{not grasp\_changeAngle\_release(\_,L,\_,\_,\_,\_,\_,T),}\\
& \qquad  \texttt{T < timemax+1.}\\
r_{m32}\ & \texttt{grasped(G,L,T+1) :- grasped(G,L,T), link(L),}\\
& \qquad  \texttt{not changeAngle\_release(L,\_,\_,G,\_,\_,\_,T),}\\
& \qquad  \texttt{not changeAngle\_release(\_,L,\_,\_,G,\_,\_,T),}\\
& \qquad  \texttt{not grasp\_changeAngle\_release(L,\_,\_,\_,\_,G,\_,T),}\\
& \qquad  \texttt{not grasp\_changeAngle\_release(\_,L,\_,\_,\_,\_,G,T),}\\
& \qquad  \texttt{T < timemax+1.}\\
r_{m33}\ & \texttt{hasAngle(L,A,T+1) :- affected(L,A,\_,T), T < timemax+1.}\\
r_{m34}\ &\texttt{hasAngle(L,A,T+1) :- hasAngle(L,A,T),}\\
& \qquad  \texttt{not changeAngle\_release(L,\_,\_,\_,\_,\_,\_,T),}\\
& \qquad  \texttt{not grasp\_changeAngle\_release(L,\_,\_,\_,\_,\_,\_,T),}\\
& \qquad  \texttt{not affected(L,\_,\_,T), T < timemax+1.}\\\\
r_{m35}\ & \texttt{\footnotesize{:- goal(J,A), not hasAngle(J,A,timemax).}}
\end{array}
$$
\caption{Macro Action Extended Scenario (MAES) encoding (Part 2).}
\label{fig:extended_tb}
\end{figure}
Figures~\ref{fig:extended_ta} and~\ref{fig:extended_tb} report the MAES encoding.
Rules $m_{1}$, $m_{2}$ and $m_{3}$ are related to the selection of possible macro actions in this model: 
$m_{1}$ is the macro that locates the joint that has to be moved to the centre of the manipulation space and then grasp two links, $m_{2}$ selects at which orientation pre-grasped links have to be moved and then release them. Rule $m_{3}$ selects two links to grasp, the orientation at which one of the link must be moved and, finally, it states that those two link must be released. \\
Moreover, $r_{m1}$ to $r_{m5}$ are used to state, respectively, which joint will be in the centre of the workspace at \texttt{T+1} time step, which links are in the robots hand and which gripper grasped which link.
Rules $r_{m6}$ to $r_{m10}$  are used to ensure that the selected angle is one of the possible angle and that the angle is always between $0$ and $360$ degrees.
Instead, rules $r_{m10}$ and $r_{m11}$  are used to compute which links are affected by the movement of the link selected by $m_{2}$, whereas, rules $r_{m13}$ update the orientation of the selected link, and rules $r_{m14}$ and $r_{m15}$ state that the grasped links are free in the next time step if $m_{2}$ is selected. 
Rules from $r_{m16}$ to $r_{m25}$ have the same purpose as rules from $r_{m6}$ to $r_{m15}$ but with respect to $m_{3}$. Then, rules $r_{m26}$ to $r_{m29}$ are used to ensure that only one macro is selected at each time step. Rules $r_{m30}$ to $r_{m33}$ are used to control the propagation overtime of the atoms \texttt{free}, \texttt{grasped}, \texttt{in\_hand} and \texttt{hasAngle} with respect to the selected action. Finally, rule $r_{m35}$ states the the goal must be reached.\\
Consider Example~\ref{ex:macro}, in which we wrote down  $\MathWord{pre}(r_{16,18})$, $\MathWord{del}(r_{16,18})$ and $\MathWord{add}(r_{16,18})$, where $r_{16,18}$ corresponds to $m_1$. Here it is possible to outline more precisely the meaning of $\MathWord{del}(r_{16,18})$ and $\MathWord{add}(r_{16,18})$.\\ Therefore, since
$$
\begin{array}{ll}
\mathit{del}(r_{16,18}) =& \{\texttt{free(G,T)}\}\\
\mathit{add}(r_{16,18}) =& \{\texttt{in\_centre(J1,T)}, \texttt{in\_hand(L1,T)}, \texttt{in\_hand(L2,T)},\\
& \texttt{grasped(G1,L1,T)}, \texttt{grasped(G2,L2,T)}\}.\\
\end{array}
$$
it is possible to notice that $\mathit{del}(r_{16,18})$ brings to the definition of $r_{m30}$ in which we state that the atom \texttt{free(G,T+1)} must not be propagated if the macro $m_1$ is selected. Moreover,  $\mathit{add}(r_{16,18})$ brings to the definition of the rules from $r_{m1}$ to $r_{m5}$ in which we state that at the next step the atoms \texttt{in\_centre(J1,T+1)}, \texttt{in\_hand(L,T+1)}, \texttt{grasped(G,L,T+1)} must be true if the macro $m_1$ is selected. 

\end{document}